\documentclass[lettersize,journal]{IEEEtran}
\usepackage{amsmath,amsfonts}
\usepackage{algorithmic}
\usepackage{algorithm}
\usepackage{array}
\usepackage[caption=false,font=normalsize,labelfont=sf,textfont=sf]{subfig}
\usepackage{textcomp}
\usepackage{stfloats}
\usepackage{url}
\usepackage{verbatim}
\usepackage{graphicx}
\usepackage{cite}

\usepackage{mathtools}
\usepackage{amsfonts}
\usepackage{amssymb}
\usepackage{bbm}
\usepackage{bm}
\usepackage{arydshln}
\usepackage{multirow}
\usepackage{booktabs}
\usepackage{pifont}
\usepackage{microtype}
\usepackage{latexsym}
\usepackage{times}
\usepackage{bbding}
\usepackage{colortbl}    
\usepackage{xcolor}      
\usepackage{CJKutf8}
\usepackage[utf8]{inputenc}
\usepackage[T1]{fontenc}
\usepackage[finnish,hungarian,english]{babel}
\usepackage{float}
\usepackage{hyperref}
\hypersetup{
    colorlinks=true,
    linkcolor=blue,
    filecolor=blue,      
    urlcolor=blue,
    citecolor=blue,
}
\usepackage{url}

\begin{document}

\renewcommand{\algorithmicrequire}{\textbf{Input:}}
\renewcommand{\algorithmicensure}{\textbf{Output:}}

\title{One Mind, Many Tongues: A Deep Dive into Language-Agnostic Knowledge Neurons in Large Language Models}


\author{
Pengfei Cao,
Yuheng Chen,
Zhuoran Jin,
Yubo Chen,
Kang Liu,
and Jun Zhao
\thanks{Pengfei Cao, Yuheng Chen, Zhuoran Jin, Yubo Chen, Kang Liu and Jun Zhao are with the Key Laboratory of Cognition and Decision Intelligence for Complex Systems, Institute of Automation, Chinese Academy of Sciences, and the School of Artificial Intelligence, University of Chinese Academy of Sciences, Beijing, China. E-mails: \{pengfei.cao, zhuoran.jin, yubo.chen, kliu, jzhao\}@nlpr.ia.ac.cn, chenyuheng22@ia.ac.cn.

}
}




\maketitle

\begin{abstract}
Large language models (LLMs) have learned vast amounts of factual knowledge through self-supervised pre-training on large-scale corpora. Meanwhile, LLMs have also demonstrated excellent multilingual capabilities, which can express the learned knowledge in multiple languages. However, the knowledge storage mechanism in LLMs still remains mysterious. Some researchers attempt to demystify the factual knowledge in LLMs from the perspective of knowledge neurons, and subsequently discover \textit{language-agnostic knowledge neurons} that store factual knowledge in a form that transcends language barriers. However, the preliminary finding suffers from two limitations: 1) \textit{High Uncertainty in Localization Results}. Existing study only uses a prompt-based probe to localize knowledge neurons for each fact, while LLMs cannot provide consistent answers for semantically equivalent queries. Thus, it leads to inaccurate localization results with high uncertainty. 2) \textit{Lack of Analysis in More Languages}. The study only analyzes language-agnostic knowledge neurons on English and Chinese data, without exploring more language families and languages. Naturally, it limits the generalizability of the findings. To address aforementioned problems, we first construct a new benchmark called Rephrased Multilingual LAMA (RML-LAMA), which contains  high-quality cloze-style multilingual parallel queries for each fact. Then, we propose a novel method named \textit{Multilingual Integrated Gradients with Uncertainty Estimation} (\textsc{MaTrice}), which quantifies the uncertainty across queries and languages during knowledge localization. Extensive experiments show that our method can accurately localize language-agnostic knowledge neurons. We also further investigate the role of language-agnostic knowledge neurons in cross-lingual knowledge editing, knowledge enhancement and new knowledge injection.
\end{abstract}


\section{Introduction}
\begin{figure}[t]
\centering
  \includegraphics[width=0.49\textwidth,height=0.5\textheight,keepaspectratio]{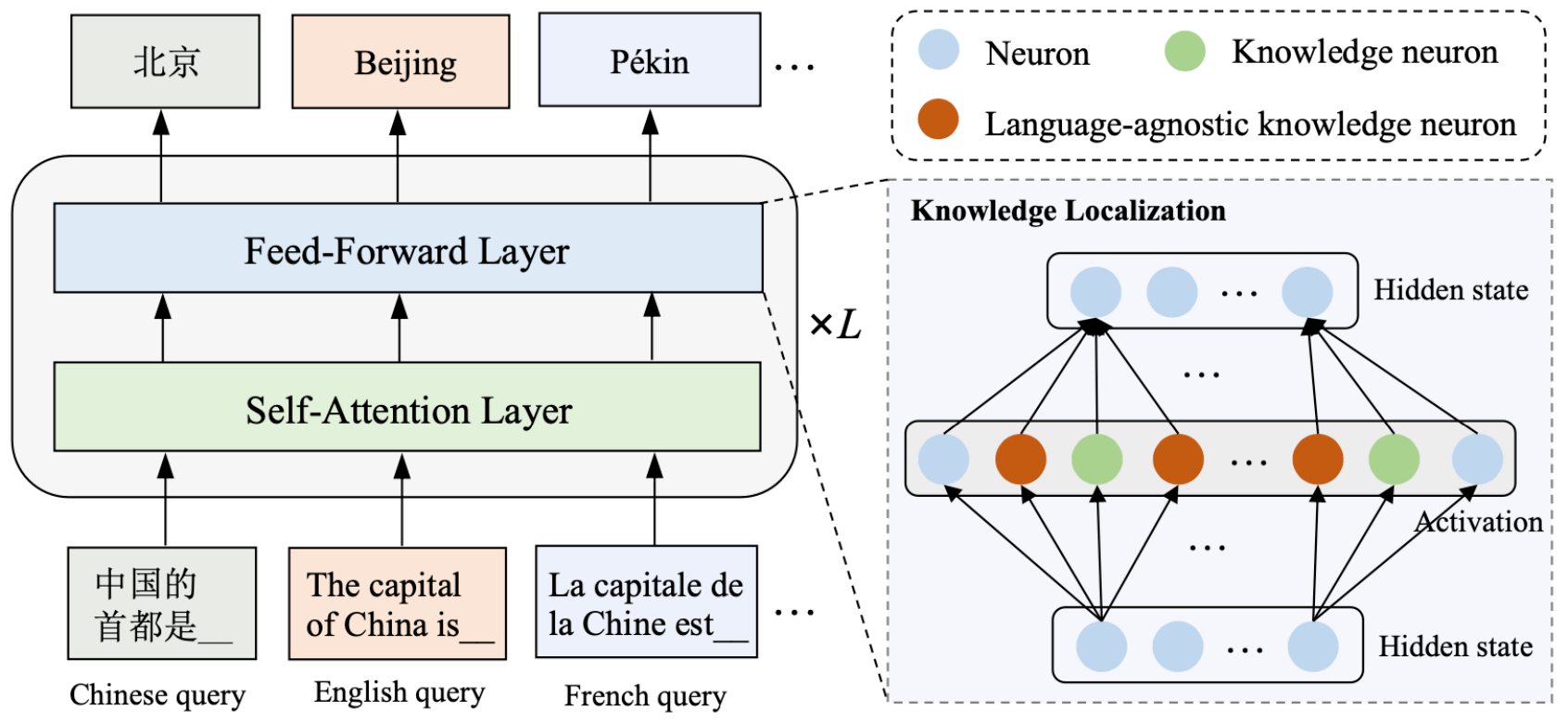}
  \caption{An example of LLMs answering multilingual queries correctly. Language-agnostic knowledge neurons are simultaneously activated by queries from multiple languages.}
  \label{fig1}
\end{figure}

\IEEEPARstart{K}{nowledge} is an important symbol of human intelligence \cite{davis1993knowledge}. How to acquire, store, represent, and apply knowledge is a critical research question in artificial intelligence \cite{mccarthy1984some,hyman1999knowledge,cao2024life}. Recently, large language models (LLMs) have achieved remarkable success and revolutionized the field of natural language processing. Through self-supervised pre-training on large-scale corpora such as Wikipedia, LLMs have learned vast amounts of factual knowledge and stored the knowledge in model parameters \cite{touvron2023llama,achiam2023gpt}. Thus, LLMs are often regarded as parametric knowledge bases\footnote{Correspondingly, knowledge graphs are seen as symbolic knowledge bases.}, which express the learned knowledge through the generation of free texts \cite{petroni2019language}. Meanwhile, LLMs have also demonstrated excellent multilingual capabilities, and are capable of expressing the same knowledge in multiple languages \cite{pires2019multilingual,shliazhko2022mgpt,qin2024multilingual}. Take Figure \ref{fig1} as an example, LLMs can correctly predict answers for queries about the fact \texttt{$\langle$China, capital, Beijing$\rangle$} expressed in Chinese, English, French, and other languages. This intuitively demonstrates the outstanding knowledge and multilingual capability of LLMs.

\begin{figure}[t]
\centering
  \includegraphics[width=0.48\textwidth,height=0.5\textheight,keepaspectratio]{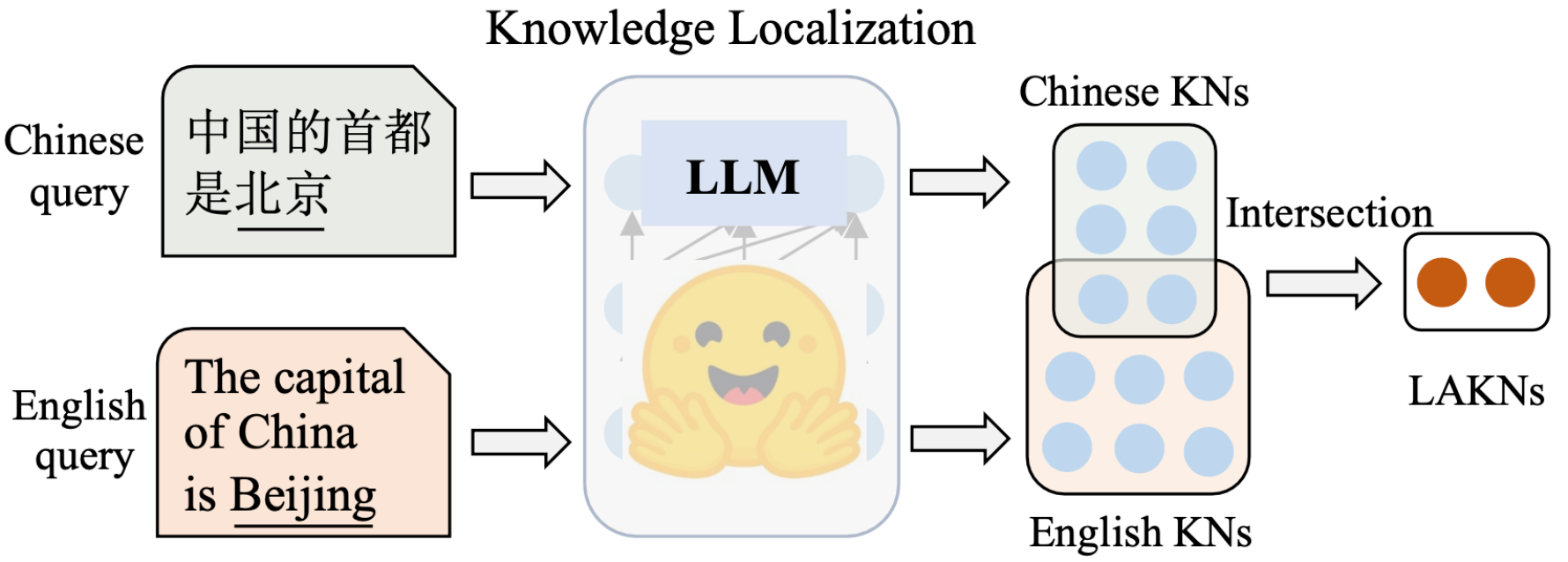}
  \caption{The process of obtaining language-agnostic knowledge neurons (LAKNs) using the previous method \cite{chen2024journey}: intersection of knowledge neurons (KNs) in Chinese and English.}
  \label{amig}
\end{figure}

Given the excellent knowledge capability, some researchers have constructed some evaluation benchmarks and attempted to use probing-based methods to assess the mastery of knowledge in LLMs \cite{petroni2019language,elazar2021measuring,yu2023kola}. To gain a deeper understanding of LLMs, some studies investigate how factual knowledge is stored in parameters \cite{dai2022knowledge,meng2022locating,chen2024journey}. For example, Dai et al.\cite{dai2022knowledge} focus on the \textit{knowledge localization} task, which aims to determine the storage location of specific factual knowledge in LLMs. They introduce a new type of neuron called \textit{knowledge neurons} that express a relational fact in feed-forward networks (FFN). Based on it, Chen et al.\cite{chen2024journey} further discover \textit{language-agnostic knowledge neurons}, which store factual knowledge in a form that transcends language barriers, as shown in Figure \ref{fig1}. Although this study fills gaps in multilingual knowledge storage mechanisms and improves interpretability of LLMs, it still suffers from two limitations: 1) \textbf{High Uncertainty in Localization Results}. As shown in Figure \ref{amig}, for each fact, the study only uses a query to locate knowledge neurons for one language. However, the LLM is highly sensitive to input \cite{elazar2021measuring,zhu2023promptbench}, meaning that it may provide different answers for rephrased queries with same semantics, which naturally leads to inconsistent localization results of knowledge neurons. As a consequence, it is highly uncertain to directly take the intersection of knowledge neurons from different languages for obtaining language-agnostic knowledge neurons. It inevitably affects the stability and reliability of the localization results. 2) \textbf{Lack of Analysis in More Languages}. The study only leverages English and Chinese data to analyze language-agnostic knowledge neurons, without exploring more language families and languages, which limits the generalizability of the findings. Moreover, LLMs do not perform equally well in all languages. In other words,  they perform better in high resource languages, but worse in low resource languages \cite{tang2024language}. However, the method overlooks the fact, which may result in inaccurate localization. Therefore, to rigorously and deeply explore multilingual knowledge storage mechanisms, we need to consider the uncertainty of localization results and mastery of each language by LLMs.

In the light of the above restrictions, we first construct a new cloze-style multilingual benchmark called  Rephrased Multilingual LAMA (\texttt{RML-LAMA}). It contains 4 language families, 7 languages, a total of 7,849 facts for 8 relations, more than 170,000 multilingual queries. Based on the dataset, we then propose a novel language-agnostic knowledge neuron localization method, termed \textbf{M}ultilingu\textbf{A}l In\textbf{T}egrated G\textbf{R}adients w\textbf{I}th Un\textbf{C}ertainty \textbf{E}stimation (\textbf{\textsc{MaTrice}}). Concretely, we first devise a \textit{Knowledge Attribution} module. For one query expressed in any language, the module leverages the gradient-based technique to compute the contribution (i.e., attribution scores) of each neuron for knowledge prediction. To calculate more accurately, we adaptively design baseline vectors for LLMs with different architecture (i.e., auto-encoding and auto-regressive models). Then, we propose an \textit{Uncertainty Quantification} module to measure the uncertainty of localization results and mastery of each language. For each neuron, the module quantifies the uncertainty of knowledge attribution across queries and languages for obtaining language-agnostic attribution score. Finally, we introduce a \textit{Knowledge Neurons Selection} module to select language-agnostic knowledge neurons by setting dynamic thresholds for different facts. 

Extensive experimental results show that a significant proportion of language-agnostic knowledge neurons are distributed in last few layers. We perform suppression and enhancement operations on language-agnostic knowledge neurons, and find that the manipulation notably affects the expression of the corresponding knowledge in various languages, which demonstrates the accuracy of our method's localization. Furthermore, we explore the potential roles of language-agnostic knowledge neurons in three aspects: 1) \textit{Cross-lingual knowledge editing}: we conduct a single edit on language-agnostic knowledge neurons can simultaneously modify the corresponding knowledge in all languages. 2) \textit{Knowledge enhancement}: by amplifying the activation values of identified language-agnostic knowledge neurons, we can improve the prediction performance of LLMs on low resource languages. 3) \textit{Knowledge injection}: we use new factual knowledge data to finetune LLMs, freezing all parameters except for those associated with language-agnostic knowledge neurons. The results indicate that language-agnostic knowledge neurons can effectively help LLMs learn new knowledge and alleviate the forgetting of old knowledge. 

Overall, our contributions are summarized as follows:

\begin{itemize}
	\item We construct a high-quality cloze-style multilingual benchmark resource called \texttt{RML-LAMA} for exploring knowledge storage mechanism in LLMs, which allows for simultaneous consideration of semantically equivalent queries, as well as different language expressions for the same fact.
 
	
	\item We propose a novel method named multilingual integrated gradients with uncertainty estimation (\textsc{MaTrice}) for the localization of language-agnostic knowledge neurons. To our best knowledge, we are the first to quantify the uncertainty across queries and languages for knowledge localization.
	
	\item Experimental results indicate that our approach can accurately localize language-agnostic knowledge neurons. We also conduct extensive experiments to analyze the properties of language-agnostic knowledge neurons and find that they are conducive to cross-lingual knowledge editing, knowledge enhancement and knowledge injection.
\end{itemize}

\section{Related Work}
In this section, we briefly review three related topics: knowledge probing for LLMs ($\S$\ref{KP}), knowledge localization for LLMs ($\S$\ref{KL}) and gradient attribution methods ($\S$\ref{GA}).

\subsection{Knowledge Probing for LLMs} \label{KP}
Knowledge probing is a prerequisite for knowledge localization, which aims to evaluate whether LLMs contain multiple types of knowledge. Prompt-based probing is the most representative methods for knowledge probing \cite{kamoda2023test,yu2023kola,wang2023readprompt}. For example, to evaluate whether LLMs know the knowledge about ``the capital of China'', we could probe LLMs with a cloze-style query ``\textit{The capital of China is} \_\_''. The prompt-based probing studies mainly focus on how to design effective prompts, which can be categorized into two types: discrete prompts and continuous prompts. For discrete prompts, Petroni et al.\cite{petroni2019language} construct the widely used LAMA benchmark, and utilize a discrete prompt-based probing method to determine whether LLMs have learned specific facts. To assess the knowledge in multilingual LLMs, Jiang et al.\cite{jiang2020x} develop a multilingual cloze-style probing benchmark, which expands the probing method from single-word to multi-word entities. Similarly, Kassner et al.\cite{kassner2021multilingual} adapt the TREx \cite{elsahar2018t} and GoogleRE\footnote{\url{https://code.google.com/archive/p/relation-extraction-corpus/}} benchmarks into a new multilingual version called mLAMA, covering multiple languages. Their experiments reveal that using mBERT \cite{devlin2018bert} as a knowledge base results in variable performance across different languages. Elazar et al.\cite{elazar2021measuring} investigate the consistency of pretrained language models in handling factual knowledge. They construct an English cloze-style benchmark called PARAREL, and their experiments indicate that the consistency of these models is lacking. Recently, Du et al.\cite{du2024zhujiu} introduce a benchmark called ZhuJiu-Knowledge to evaluate commonsense knowledge, factual knowledge and linguistic knowledge for LLMs. To guide LLMs towards more accurate predictions, some works optimize the prompts in a continuous space \cite{liu2023gpt,zhong2021factual,li2021prefix}. In that way, the prompt is composed of trainable vectors that are not necessarily related to actual words. Although the continuous prompts may achieve better results, they compromise the readability to humans\footnote{To ensure readability, we use discrete prompts for knowledge localization in this paper.} \cite{khashabi2022prompt}.

\subsection{Knowledge Localization for LLMs} \label{KL}
To investigate knowledge storage mechanisms in LLMs, some researchers attempt to study the knowledge localization to determine the storage location of specific factual knowledge \cite{dai2022knowledge,wang2024knowledge}. For example, Geva et al.\cite{geva2021transformer} hypothesize that the factual knowledge is stored in FFN layers, and propose that the function of FFN layers in LLMs is similar to key-value memory, where the first layer represents keys and the second layer corresponds to values. Based on the empirical findings, Dai et al.\cite{dai2022knowledge} introduce the concept of knowledge neurons, and propose a knowledge attribution method to identify the neurons with higher attribution scores. Meng et al.\cite{meng2022locating,meng2022mass} utilize a causal intervention method to find knowledge neurons that are decisive in predicting the factual knowledge. To verify the effectiveness of localization, they modify these knowledge neurons to observe whether the corresponding factual knowledge is updated. Chen et al.\cite{chen2024vinci} conduct further exploration on knowledge neurons, and discover a new kind of neurons called degenerate knowledge neurons, which denotes that some subsets of knowledge neurons can independently express the same fact. It is similar to the degenerate phenomenon in biological systems \cite{tononi1999measures,mason2015degeneracy}. Furthermore, given a subject-relation prompt query, Geva et al.\cite{geva2023dissecting} strive to figure out the information flow process of LLMs from aggregating information from the subject and relation to predicting the corresponding attribute. Similar to our work, Chen et al.\cite{chen2024journey} leverage the gradient-based method to localize knowledge neurons. Based on it, they discover language-agnostic knowledge neurons that store factual knowledge across languages. However, for one fact, they only use a prompt-based query to localize knowledge neurons, but the predictive performance of LLMs varies greatly for different queries \cite{elazar2021measuring,chen2024knowledge}. It may result in inaccurate localization results. Besides, they only use two languages (i.e., Chinese and English) to identify language-agnostic knowledge neurons without considering other languages, thus limiting the generalizability of the findings to some extent. Therefore, it is necessary to consider multiple queries expressed in multiple languages for one fact in knowledge localization tasks.

\subsection{Gradient Attribution Methods} \label{GA}
To improve the interpretability of deep learning models, various methods have been proposed to attribute the model’s output to specific input features. A widely used baseline is the product of gradients and feature values \cite{baehrens2010explain,simonyan2013deep}. For example, Sundararajan et al.\cite{sundararajan2017axiomatic} present the axiomatic attribution method, emphasizing sensitivity and implementation invariance as crucial principles for attribution. This approach results in the development of integrated gradients. Subsequently, Sanyal et al.\cite{sanyal2021discretized} introduce discretized integrated gradients to explain models that process discrete text data. They develop two interpolation strategies to create non-linear interpolation paths in the word embedding space. Enguehard et al.\cite{enguehard2023sequential} propose the sequential integrated gradients technique, which assesses the significance of individual words within a sentence by holding all other words constant and interpolating solely between the baseline and the word of interest. Liu et al.\cite{liu2022effective} introduce a method called shapley integrated gradients that constructs baseline vectors by employing proportional sampling to approximate the computation pathway of shapley values. For gradient attribution methods, one of main challenges is how to design effective baseline vectors. Our method for knowledge localization is built upon integrated gradients, and improve the baseline vectors to minimize their information content.

\begin{figure*}
	\centering
	\includegraphics[width=0.88\textwidth,height=0.6\textheight,keepaspectratio]{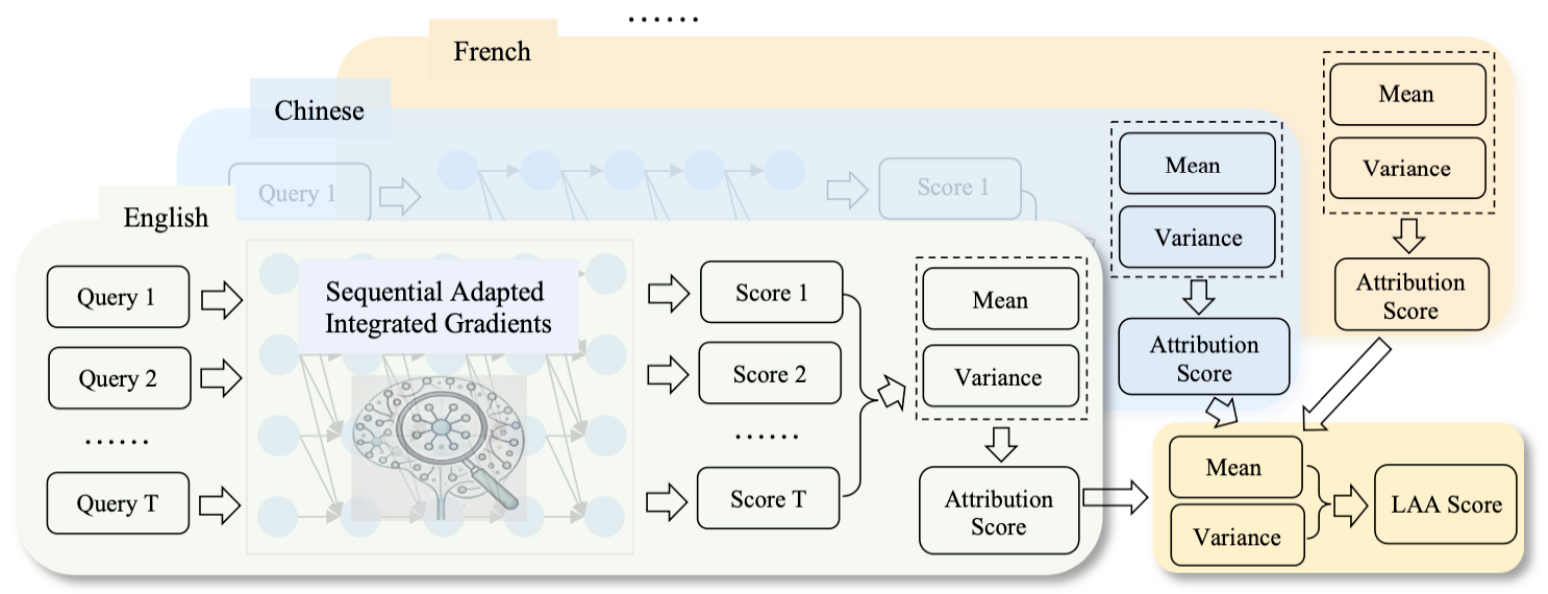}
	\caption{The architecture of our proposed multilingual integrated gradients with uncertainty estimation (\textsc{MaTrice}) for calculating the language-agnostic attribution score for each neuron. ``Query 1'', ``Query 2'' and ``Query T'' denote semantically equivalent queries for one fact. ``Score 1'' is the attribution score for each neuron, when using ``Query 1'' for knowledge attribution. ``LAA score'' denotes the language-agnostic attribution score.} 
	\label{fig2}
\end{figure*}

\section{Method}
Figure \ref{fig2} schematically visualizes our proposed method, which consists of three major components: (1) \textit{Knowledge Attribution} ($\S$\ref{KA}), which preliminarily computes the attribute score for each neuron using the gradient-based technique; (2) \textit{Uncertainty Quantification} ($\S$\ref{UQ}), which estimates the uncertainty of knowledge attribution across queries and languages to compute language-agnostic attribution score for each neuron; and (3) \textit{Knowledge Neurons Selection} ($\S$\ref{KNS}), which devises an adaptive threshold to select language-agnostic knowledge neurons. We will illustrate each component in detail.

\subsection{Knowledge Attribution} \label{KA}
Following previous studies \cite{dai2022knowledge,chen2024journey}, we utilize the fill-in-the-blank cloze task to evaluate whether a LLM knows a specific fact. That is to say, given a fact, we test if LLMs can answer the cloze query that expresses the fact but leaves the tail entity as a blank. For example, given a fact \texttt{$\langle$China, capital, Beijing$\rangle$} with a possible query ``\textit{The capital of China is} \_\_'', we assume that a LLM has mastered the fact if it can predict the correct answer. Based on it, we propose the Sequential Adapted Integrated Gradients (SAIG) technique to assess the importance of each neuron for expressing factual knowledge.

Formally, given a query $q$,  the probability of the correct answer predicted by a LLM is computed as follows:
\begin{equation}
\label{E1}
{\rm P}(\bm{w}_{i}^{l})=p(y^{*}|q, n_{i}^{l}=\bm{w}_{i}^{l}),
\end{equation}
where $y^{*}$ is the correct answer for the query $q$. $n_{i}^{l}$ denotes the $i$-th neuron in the $l$-th FFN layer of the LLM. $\bm{w}_{i}^{l}$ is the value assigned to the neuron $n_{i}^{l}$. To measure the contribution of each neuron for prediction, the integrated gradients technique \cite{sundararajan2017axiomatic} can be used to compute the attribution score. For the neuron $n_{i}^{l}$, its attribution score is calculated as follows:
\begin{equation}
\label{E2}
\begin{aligned}
{\rm Attr}(n_{i}^{l})=&\Delta \bm{w}_i^{l}\int_{\alpha=0}^{1}\frac{\partial {\rm P}(\bm{w}_{i}^{\prime l}+\alpha \Delta \bm{w}_i^{l})}{\partial \Bar{\bm{w}}_i^{l}}d\alpha, \\
&\Delta \bm{w}_i^{l}=\Bar{\bm{w}}_i^{l} - \bm{w}^{\prime l}_i,
\end{aligned}
\end{equation}
where $\frac{\partial {\rm P}(\bm{w}_{i}^{\prime l}+\alpha \Delta \bm{w}_i^{l})}{\partial \Bar{\bm{w}}_i^{l}}$ computes the gradient of the probability for the correct answer with regard to $\Bar{\bm{w}}_i^{l}$. $\Bar{\bm{w}}_{i}^{l}$ is the value of the neuron $n_{i}^{l}$, and $\bm{w}^{\prime l}_i$ is the corresponding baseline vector. $\alpha$ controls the integration from the baseline vector $\bm{w}^{\prime l}_i$ to the original value $\Bar{\bm{w}}_{i}^{l}$ of the neuron $n_{i}^{l}$ by changing from 0 to 1. In this way, ${\rm Attr}(n_{i}^{l})$ enables to accumulate the output probability change. Naturally, if a neuron significantly impacts the expression of a fact, it will have a prominent gradient, leading to high integration values. Thus, the attribution score can quantify how much the neuron contributes to the expression of the fact.

For convenience of calculations, previous studies simply set the baseline vector $\bm{w}^{\prime l}_i$ to zero vector \cite{dai2022knowledge}. However, neurons with zero activation values still contain information, thus directly using the zero vector as the baseline vector is not the optimal solution to measure the importance of each neuron in knowledge expression \cite{chen2024journey}. To this end, for the query $q=\{x_{1}, x_{2}, \ldots, x_{m}\}$, consisting of $m$ tokens, we replace a token with a special token \texttt{$\langle$PAD$\rangle$} at each time to construct its corresponding baseline sentence. The special token is defined as follows:
\begin{equation}
  \langle {\rm PAD}\rangle = 
  \begin{cases} 
      \langle {\rm mask} \rangle, & \text{for auto-encoding models}, \\
      \langle {\rm eos} \rangle,  & \text{for auto-regressive models}, 
  \end{cases}
\end{equation}
where $\langle {\rm mask} \rangle$ is used for masking auto-encoding models (e.g., BERT), $\langle {\rm eos} \rangle$ is the end-of-sequence token in auto-regressive models (e.g., GPT). Since $\langle {\rm mask} \rangle$ or $\langle {\rm eos} \rangle$ is trained to replace random tokens, therefore the special token does not have a specific meaning, and  seems more suited to be the baseline. For the token $x_{i}$, its corresponding baseline sentence is denoted as $s_{i}=\{x_{1},\ldots, x_{i-1}, \langle {\rm PAD}\rangle, x_{i+1}, \ldots, x_{m}\}$. By traversing each token in this way, the set of baseline sentences can be obtained, denoted as $Q=\{s_{1}, s_{2}, \ldots, s_{m}\}$. We feed each baseline sentence separately into the LLM to calculate the sentence embeddings as baseline vectors $V=\{\bm{w}_{1}^{\prime}, \bm{w}_{2}^{\prime}, \ldots, \bm{w}_{m}^{\prime}\}$. When using $\bm{w}^{\prime}_i$ as the baseline vector, the attribution score of the neuron $n_{j}^{l}$ is denoted as ${\rm Attr}_{i}(n_{j}^{l})$. Since directly calculating continuous integrals is intractable, we instead use the Riemann approximation of the integration to compute the attribution score as follows:
\begin{equation}
\label{E4}
\begin{aligned}
    {{\rm Attr}_{i}(n_j^{l})}  \approx  & \frac{\Delta \bm{w}_j^{l}}{M} \sum_{k=1}^{M} \frac{ \partial {\rm P}(\bm{w}_{i}^{\prime} + \frac{k}{M} \times \Delta \bm{w}_j^{l})}{\partial \Bar{\bm{w}}_j^{l}}, \\
&\Delta \bm{w}_j^{l}=\Bar{\bm{w}}_j^{l} - \bm{w}^{\prime}_i,
\end{aligned}
\end{equation}
where $M$ is the number of approximation steps. The attribution score for each token is summed and then normalized to obtain the final attribution score for each neuron:
\begin{equation}
\label{E_5}
{{\rm Attr}(n_j^{l})} = \frac{\sum_{i=1}^{m}{\rm Attr}_{i}(n_j^{l})}{\sum_{j=1}^{N}\sum_{i=1}^{m}{\rm Attr}_{i}(n_j^{l})},
\end{equation}
where $N$ is the number of neurons in the $l$-th layer.

\subsection{Uncertainty Quantification} \label{UQ}
For one fact, it can be expressed by multiple different rephrased queries in the same language, and the semantically equivalent queries are denoted as $\{q_{1}, q_{2}, \ldots, q_{T}\}$, where $T$ is the number of queries. Given the query $q_{i}$, we can use the above described SAIG technique to obtain the attribution score of each neuron for knowledge prediction, and the attribution score for the $i$-th neuron of $l$-th layer is denoted as ${\rm Attr}(n_i^{l})$. Since LLMs are highly sensitive to queries and have varying levels of proficiency in different languages, we need to consider the uncertainty of knowledge attribution across queries and languages.

\paragraph{Uncertainty across Queries}
\label{method_a}
We first compute the attribution score of each neuron (e.g., $n_i^{l}$) using multiple semantically equivalent queries $\{q_{1}, q_{2}, \ldots, q_{T}\}$, denoted as $\{[{\rm Attr}(n_i^{l})]_{1}, [{\rm Attr}(n_i^{l})]_{2}, \ldots, [{\rm Attr}(n_i^{l})]_{T}\}$. Then, we quantify the uncertainty of attribution scores for these queries. Here, we consider the uncertainty in terms of the expectation and variance. The variance of ${\rm Attr}(n_i^{l})$ can be approximated as follows:
\begin{equation}
\label{E_6}
\begin{split}
{\rm Var}[{\rm Attr}(n_i^{l})] & = \mathbb{E}[({\rm Attr}(n_i^{l})-\mathbb{E}[{\rm Attr}(n_i^{l})])^{2}] \\
& = \frac{1}{T}\sum_{j=1}^{T}([{\rm Attr}(n_i^{l})]_{j}-\mathbb{E}[{\rm Attr}(n_i^{l})])^{2},
\end{split}
\end{equation}
where $[{\rm Attr}(n_i^{l})]_{j}$ represents the attribute score of the neuron $n_i^{l}$ when querying LLMs using $q_{j}$. $\mathbb{E}[{\rm Attr}(n_i^{l})]$ denotes the expectation of ${\rm Attr}(n_i^{l})$ that can be computed as follows:
\begin{equation}
\label{E_7}
\mathbb{E}[{\rm Attr}(n_i^{l})] = \frac{1}{T}\sum_{j=1}^{T}[{\rm Attr}(n_i^{l})]_{j}.
\end{equation}

Intuitively, the more important the neuron is for knowledge expression, the higher expectation and lower variance of attribution score are. Therefore, we combine the mean and variance to obtain a more accurate attribution score $s_{i}^{l}$ for neuron $n_i^{l}$ as follows:
\begin{equation}
\label{E8}
s_{i}^{l}=\alpha_{1}\mathbb{E}[{\rm Attr}(n_i^{l})]-\alpha_{2}\sqrt{{\rm Var}[{\rm Attr}(n_i^{l})]},
\end{equation}
where $\alpha_{1}$ and $\alpha_{2}$ are hyperparameters. In this way, we can obtain more accurate attribution scores for each neuron by quantifying the uncertainty across paraphrased queries expressed in a certain language.

\begin{algorithm}[t]
	\caption{The Procedure of Obtaining LAKNs} 	
	\label{alg:Framwork} 
	\begin{algorithmic}[1]
  \REQUIRE  Large language model $\mathcal{M}$, facts set $\mathcal{E}$ with
  multilingual paraphrase queries $\mathcal{D}=\{\mathcal{D}_{1}, \mathcal{D}_{2}, \ldots, \mathcal{D}_{L}\}$, hyperparameters $\alpha$, $\alpha_{1}$, $\alpha_{2}$, $\beta_{1}$, $\beta_{2}$ and $\tau$.  \\
  \ENSURE Language-agnostic knowledge neurons $\mathcal{N}$. \\
      \FOR {language $l$ in $\{\mathcal{L}_{1}, \mathcal{L}_{2}, \ldots, \mathcal{L}_{L}\}$}
		\FOR {query $q$ in $\mathcal{D}_{l}$}
		\STATE Compute the predicted probability of the correct answer by the LLM via Equation (\ref{E1}) \\
		\STATE Approximately compute the attribution score ${\rm Attr}(\cdot)$ for each neuron via Equation (\ref{E_5}) \\
		\ENDFOR
      \ENDFOR
     \FOR {each fact $e$ in $\mathcal{E}$}
        \STATE Quantify the uncertainty (i.e., the variance ${\rm Var}[{\rm Attr}(\cdot)]$ and the mean $\mathbb{E}[{\rm Attr}(\cdot)]$) across queries via Equation (\ref{E_6}) and (\ref{E_7})
        \STATE Combine the mean and variance to obtain a more accurate attribution score $s$ via Equation (\ref{E8})
		\STATE Normalize the attribution score for each language via Equation (\ref{E9})
         \STATE Measure the uncertainty (i.e., the variance ${\rm Var}[\Bar{s}_{i}^{l}]$ and the mean $\mathbb{E}[\Bar{s}_{i}^{l}]$) across languages via Equation (\ref{E10})
		\STATE Compute the language-agnostic activation score via Equation (\ref{E11})	
		\STATE Select the language-agnostic knowledge neurons $\mathcal{N}$ based on a dynamic threshold $\mathcal{T}$ via Equation (\ref{E13})
    \ENDFOR
	\end{algorithmic}	
\end{algorithm}

\paragraph{Uncertainty across Languages}
\label{method_b}
On the other hand, since knowledge expression is independent of languages, the same fact can be expressed in multiple languages. For one fact represented by language $k$, we can obtain the attribution score $[s_{i}^{l}]_{k}$ for the neuron $n_{i}^{l}$ via the Equation (\ref{E8}). As a result, the attribution scores of the neuron $n_{i}^{l}$ for multiple languages are denoted as $\{[s_{i}^{l}]_{1}, [s_{i}^{l}]_{2}, \ldots, [s_{i}^{l}]_{L}\}$, where $L$ is the number of languages. Although LLMs have multilingual abilities, their mastery of different languages varies. Therefore, for the same fact, the responses of a LLM to queries in different languages are different, resulting in significant differences in attribution values for different languages. To this end, we first normalize the attribution scores as follows:
\begin{equation}
\label{E9}
[\Bar{s}_{i}^{l}]_{j} = \frac{[s_{i}^{l}]_{j} - \min_{k}([s_{i}^{l}]_{k})}{\max_{k}([s_{i}^{l}]_{k}) - \min_{k}([s_{i}^{l}]_{k})}, \quad j, k \in \{1, 2, \ldots, L\}.
\end{equation}

Then, we delve into the uncertainty across languages to obtain language-agnostic attribution scores for each neuron. Similar to uncertainty quantification across queries, we compute the mean and variance of the normalized attribution scores across languages to assess the uncertainty as follows:
\begin{equation}
\label{E10}
\begin{aligned}
    &\mathbb{E}[\Bar{s}_{i}^{l}] = \frac{1}{L}\sum_{j=1}^{L}[\Bar{s}_{i}^{l}]_{j}, \\
{\rm Var}[\Bar{s}_{i}^{l}]&=\frac{1}{L}\sum_{j=1}^{L}([\Bar{s}_{i}^{l}]_{j}-\mathbb{E}[\Bar{s}_{i}^{l}])^{2}.
\end{aligned}
\end{equation}

Based on the above mean and variance, we compute the language-agnostic attribution score for each neuron as follows:
\begin{equation}
\label{E11}
S_{i}^{l}=\beta_{1}\mathbb{E}[\Bar{s}_{i}^{l}]-\beta_{2}\sqrt{{\rm Var}[\Bar{s}_{i}^{l}]},
\end{equation}
where $\beta_{1}$ and $\beta_{2}$ are hyperparameters. $S_{i}^{l}$ is the language-agnostic attribution score for the neuron $n_{i}^{l}$.

\subsection{Knowledge Neurons Selection} \label{KNS}
Finally, we select neurons based on language-agnostic attribution scores. Since the level of mastery of different facts varies in large language models, it is difficult to obtain language-agnostic knowledge neurons for all facts by setting a unified threshold. We set the scaling factor $\tau$ for the neuron selection, and calculate the dynamic threshold for a certain fact as follows:
\begin{equation}
\label{E12}
\mathcal{T} = \max_{i, l} S_{i}^{l} \times \tau.
\end{equation}

Then, we select neurons with attribution score exceeding the threshold: 
\begin{equation}
\label{E13}
\mathcal{N}=\{n_{i}^{l} | S_{i}^{l}>\mathcal{T}, \ \forall i, l\},
\end{equation}
where $\mathcal{N}$ is the set of the language-agnostic knowledge neurons for a certain fact. The overall procedure of obtaining language-agnostic knowledge neurons is outlined in Algorithm \ref{alg:Framwork}.

\begin{figure*}[t]
\centering
\includegraphics[width=\linewidth]{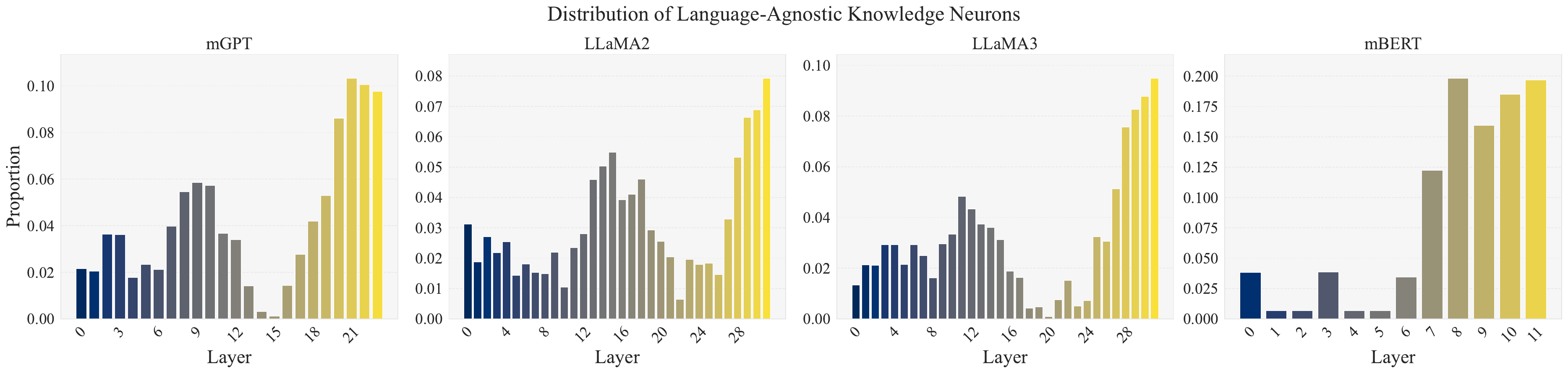} 
\caption{The distribution of language-agnostic knowledge neurons in four multilingual LLMs, including mGPT, LLaMA2, LLaMA3 and mBERT.}
    \label{fig-distribution}
\end{figure*}

\section{Experiments}
In this section, we conduct extensive experiments with the aim of answering the following research questions (\textbf{RQs}):
\begin{itemize}
  \item \textbf{RQ1}: How effective is our method \textsc{MaTrice} in obtaining language-agnostic knowledge neurons? ($\S$\ref{Manipulation})
  \item \textbf{RQ2}: How does each design of the proposed method matter? ($\S$\ref{AS})
 \item \textbf{RQ3}: What are the potential applications of language-agnostic knowledge neurons? ($\S$\ref{KE}, $\S$\ref{KELRL} and $\S$\ref{LEK})
\end{itemize}

In the remainder of this section, we describe the new constructed dataset ($\S$\ref{dataset}), the LLMs ($\S$\ref{LLMs}) that are analyzed in the experiments, the distribution of language-agnostic knowledge neurons ($\S$\ref{localization}), and the case study ($\S$\ref{case study}).

\subsection{Rephrased Multilingual LAMA} \label{dataset}
The mLAMA \cite{kassner2021multilingual} is an extension of LAMA \cite{petroni2019language}, which is used for multilingual knowledge probing. However, the dataset only provides one query for each fact. To address this, we construct a new benchmark called Rephrased Multilingual LAMA (\texttt{RML-LAMA}), in which each fact corresponds to several semantically equivalent queries in multiple languages. We use the Claude 3 Opus API\footnote{\url{https://www.anthropic.com/api}} to generate these rephrased queries. In detail, the construction process involves three steps:
1) \textit{Utilization of Existing Queries}: We utilize English rephrased queries from LAMA as prompts. 2) \textit{Translation with Fact Integration}: We combine these prompts with the corresponding fact in the target language.
For example, given an English query \textit{[X] was born in [Y]}, its rephrased query in LAMA is \textit{[X] is originally from [Y]}, its French copy in mLAMA is \textit{[X] syntyi [Y]}, then the rephrased queries in French can be generated based on these three queries. We choose low-quality generated samples, offer detailed explanations for the required corrections, and then regenerate them. 3) \textit{Query Formation}: Finally, we replace placeholders like \textit{[X]} and \textit{[Y]} with the subject and object of the fact to form complete queries.

The dataset contains 4 language families (Sino-Tibetan languages, Indo-European languages, Uralic languages, and Altaic languages), 7 languages (Chinese (zh), English (en), French (fr), Finnish (fi), Hungarian (hu), Japanese (ja) and Korean (ko)),  a total of 7,849 facts for 8 relations. The detailed statistical information of the dataset is presented in Table \ref{data}\footnote{For some facts, the difficulty of rephrasing queries varies across languages, thus the number of queries (i.e., \#Queries) corresponding to each language is different.}.

\begin{figure*}[t]
\centering
\includegraphics[width=1.0\textwidth,height=0.9\textheight,keepaspectratio]{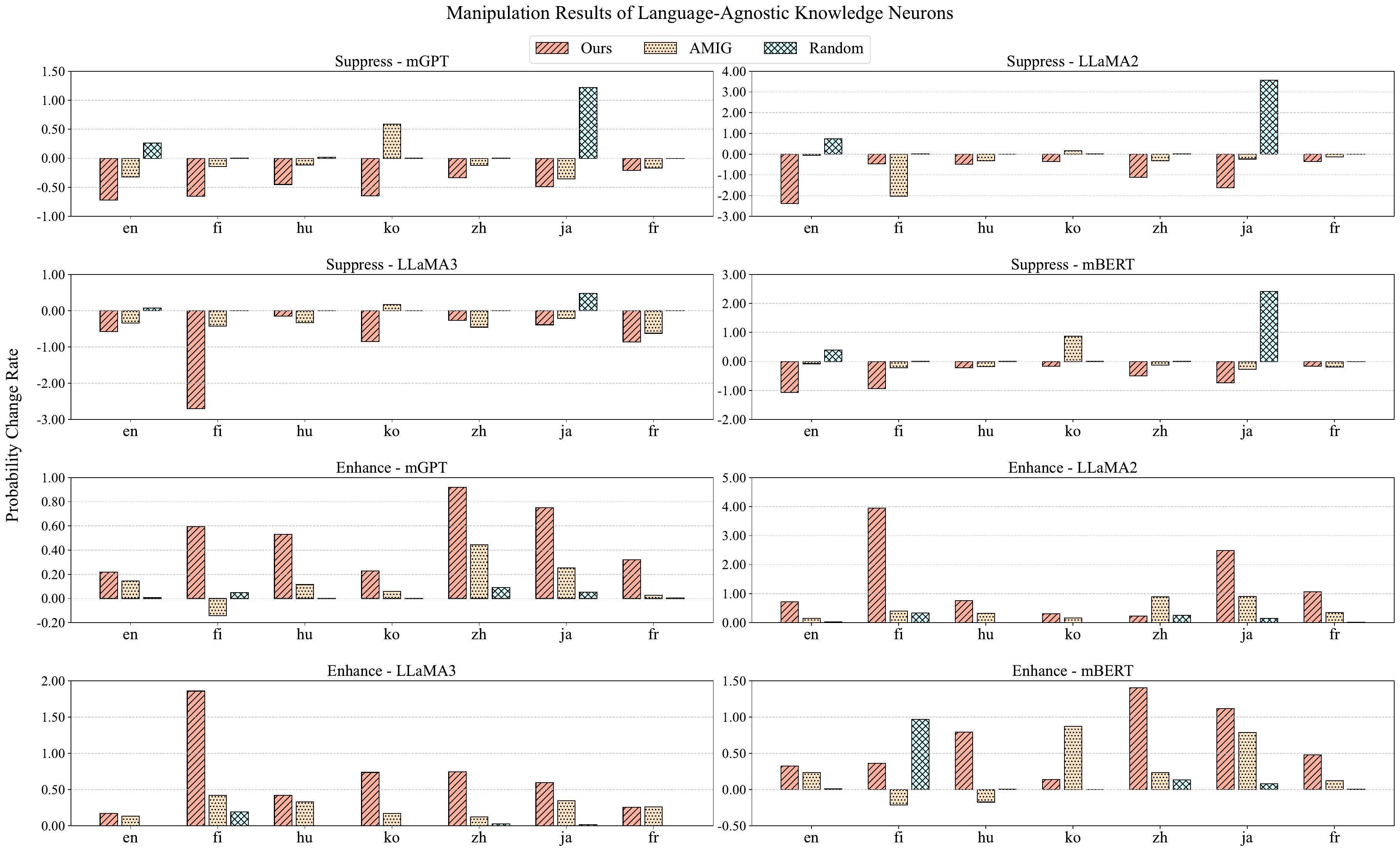} 
\caption{Results of suppressing or enhancing language-agnostic knowledge neurons experiment. The ``Probability Change Rate'' denotes the probability change ratio of correct answers, which is computed via Equation (\ref{E14}).}
    \label{fig-suppression or enhancement}
\end{figure*}

\begin{table}[t]
    \centering
    \caption{The statistical information for the rml-lama benchmark. ``Rel'' indicates ``Relations''. ``Avg.Len'' denotes the average length of the queries.}
    
    \scalebox{0.96}{
    \begin{tabular}{c|c|cccc}
    \toprule
         Language Family & Language & \#Facts & \#Queries & \#Rel & Avg.Len  \\
         \midrule
         Sino-Tibetan & Chinese & 7,849  & 23,553 & 8 & 11.2 \\
         \midrule
         \multirow{2}{*}{Indo-European} &  English & 7,849 & 31,669 & 8 & 6.3 \\
          & French & 7,849 & 26,440 & 8 & 6.8 \\
         \midrule
         \multirow{2}{*}{Uralic}  & Finnish & 7,849 & 22,782 & 8 & 5.0 \\
          & Hungarian & 7,849 & 20,432 & 8 & 4.5  \\
         \midrule
         \multirow{2}{*}{Altaic}  & Japanese & 7,849 & 23,129 & 8 & 19.4  \\
         & Korean & 7,849 & 23,736 & 8 & 17.4  \\
         \bottomrule
    \end{tabular}
    }
    \label{data}
\end{table}

\subsection{Analyzed Large Language Models} \label{LLMs}
We conduct the experiments based on two types of LLM architectures: auto-encoding model and auto-regressive model. For the auto-encoding model, we use the mBERT \cite{devlin2018bert} to analyze language-agnostic knowledge neurons. For the auto-regressive model, we opt for three popular models, including mGPT \cite{shliazhko2022mgpt}, LLaMA2-7b \cite{touvron2023llama} and LLaMA3-8b \cite{meta2024introducing}. 
Our primary focus is on auto-regressive models, and we test models with varying number of parameters to verify the scalability of our method across different model sizes. Additionally, we include an auto-encoding model to demonstrate the applicability of our method across different transformer architectures. In this way, it allows us to assess the robustness and versatility of our method across diverse model types and sizes.


\subsection{Distribution of Language-Agnostic Knowledge Neurons} \label{localization}
Figure \ref{fig-distribution} shows the layer-wise distribution of language-agnostic knowledge neurons across each layer in four multilingual LLMs, including mGPT, LLaMA2, LLaMA3 and mBERT. From the figure, we can observe the following phenomena: 1) For the four multilingual LLMs, the last few layers contain a higher proportion of language-agnostic knowledge neurons.  2) For the auto-regressive models (i.e., mGPT, LLaMA2 and LLaMA3), their language-agnostic knowledge neurons have similar distribution patterns. The shallow and middle layers also contain a certain proportion of language-agnostic knowledge neurons. However, for the auto-encoding model (i.e., mBERT), the shallow layers contain fewer language-agnostic knowledge neurons, and as the number of layers increases, the proportion of language-agnostic knowledge neurons also increases.

\subsection{Manipulation Results of Language-Agnostic Knowledge Neurons} 
\label{Manipulation}
In this subsection, we start to address the research question \textbf{RQ1} via the manipulation experiment of language-agnostic knowledge neurons. 

\textbf{Experimental Settings:} Following previous studies \cite{dai2022knowledge,chen2024journey}, given a fact, we manipulate the corresponding language-agnostic knowledge neurons to observe the probability change rate of correct answers. The manipulation operation includes: 1) \textit{suppressing} these neurons by setting their activation values to zero; 2) \textit{enhancing} these neurons by doubling their activation values. 

\textit{Baseline}: We compare our method with two representative baselines: 1) \textbf{AMIG} \cite{chen2024journey}, which obtains language-agnostic knowledge neurons by intersecting knowledge neurons corresponding to each language; 2) \textbf{Random}, which randomly selects the same number of neurons as the \textsc{MaTrice} (Ours) method to conduct the manipulation experiment.

\textit{Evaluation Metric}: We measure performance using the probability change rate ($\Delta {\rm P}$) of correct answers before and after manipulation as follows: 

\begin{equation}
\Delta {\rm P} = \frac{{\rm P}_{{\rm after}} - {\rm P}_{{\rm before}}}{{\rm P}_{{\rm before}}},
\label{E14}
\end{equation}
where ${\rm P}_{{\rm before}}$ and ${\rm P}_{{\rm after}}$ denote the probabilities of correct answers before and after manipulation operation. For suppression, $\Delta {\rm P}$ is typically negative, with lower values indicating better performance. For enhancement, higher positive $\Delta {\rm P}$ values indicate better performance.  

\textbf{Results and Analysis:} In Figure \ref{fig-suppression or enhancement}, we plot the probability change rate of the correct answer for the three methods. From the results, we have the following observations:

1) Compared with the AMIG method, when suppressing or enhancing language-agnostic knowledge neurons, our method can consistently result in negative or positive probability changes. This demonstrates that the language-agnostic knowledge neurons localized by our method are more precise. Moreover, our method achieves more significant probability change rate of the correct answer, which may be due to the limited number of language-agnostic knowledge neurons. AMIG obtains language-agnostic knowledge neurons by intersecting knowledge neurons of different languages, while as the number of languages increases, the number of obtained language-agnostic knowledge neurons will be fewer. Therefore, the influence of manipulating language-agnostic knowledge neurons obtained via the AMIG method is not obvious.

2) The non-zero effects of the Random method can be attributed to two reasons. On the one hand, the random selection process does not exclude neurons identified by the AMIG or our method. Consequently, it may include some of the most influential neurons, leading to the non-zero effects. On the other hand, some neurons may be crucial for specific languages but not others. Our method and the AMIG might exclude language-specific neurons, while these neurons can still affect performance if randomly selected. These two reasons together explain the observed effects in the random condition.

3) In rare cases, the Random method outperforms the other two methods, possibly due to fortuitous selection of highly impactful neurons or capturing language-specific effects missed by our method and the AMIG. These exceptions highlight the complex nature of neuron interactions and the potential for unexpected outcomes in neurons manipulation.

\subsection{Ablation Study} \label{AS}
To demonstrate the effectiveness of \textit{Uncertainty Quantification} module, we conduct an ablation study for answering the research question \textbf{RQ2}. 

\textbf{Experimental Settings:} We ablate the \textit{Uncertainty across Queries} (UaQ) and \textit{Uncertainty across Languages} (UaL) separately to create two variants: one without UaQ (i.e., w/o UaQ) and another without UaL (i.e., w/o UaL). Specifically, when not using UaQ, we only use one query for each fact to conduct knowledge attribution, then apply UaL to obtain language-agnostic attribution score for each neuron. Based on the score, we can select language-agnostic knowledge neurons. When not using UaL, we first obtain neurons for each language using UaQ, then take the intersection of neurons across different languages to obtain language-agnostic knowledge neurons. For the two variants and complete \textsc{MaTrice}, we perform the same manipulation operations (i.e., suppression and enhancement) on language-agnostic knowledge neurons as described in $\S$\ref{Manipulation}, and observe output probability changes of LLMs.

\textbf{Results and Analysis:} Table \ref{tab:ablation_results} presents the results of ablation study. To clearly illustrate the effect of each component, we normalize the results for each language separately and then average them across languages. The normalization process involves setting our complete method (\textsc{MaTrice}) as the baseline (1.000) for both enhancement  and suppression operations across the four LLMs. For the  two variants, their performance is presented as a fraction of \textsc{MaTrice}'s performance. For instance, a value of 0.830 indicates that the method achieves 83.0\% of \textsc{MaTrice}'s performance, representing a 17.0\% decrease in effectiveness. From the results, we can see that removing any module will result in significant performance degradation, which demonstrates that both components contribute to the effectiveness of our approach. Besides, with the absence of UaQ generally having a more pronounced impact on performance compared to the absence of UaL. For example, removing UaQ will result in an average performance of 84.0\% of the complete model on the mGPT, while removing UaL will result in an average performance of 91.5\%. It suggests that while both components are important, the UaQ plays a particularly crucial role in our method's success.

\begin{table}
\centering
\caption{Ablation study by removing the main components, where ``w/o'' indicates without. ``Average'' is the average value of the results via the enhancement and suppression operations for language-agnostic knowledge neurons.}
\scalebox{1}{
\begin{tabular}{ccccc}
\toprule
Base Models             & Methods & Enhance & Suppress & Average \\
\midrule
\multirow{3}{*}{mGPT}   & \textsc{MaTrice} & 1.000 & 1.000 & 1.000    \\
                        & \, w/o UaQ & 0.830 & 0.850 & 0.840    \\
                        & \, w/o UaL & 0.910 & 0.920 & 0.915   \\
\midrule
\multirow{3}{*}{LLaMA2} & \textsc{MaTrice} & 1.000 & 1.000 & 1.000    \\
                        & \, w/o UaQ & 0.870 & 0.860 & 0.865   \\
                        & \, w/o UaL & 0.930 & 0.940 & 0.935   \\
\midrule
\multirow{3}{*}{LLaMA3} & \textsc{MaTrice} & 1.000 & 1.000 & 1.000    \\
                        & \, w/o UaQ & 0.850 & 0.840 & 0.845   \\
                        & \, w/o UaL & 0.920 & 0.930 & 0.925   \\
\midrule
\multirow{3}{*}{mBERT}  & \textsc{MaTrice} & 1.000 & 1.000 & 1.000    \\
                        & \, w/o UaQ & 0.820 & 0.830 & 0.825   \\
                        & \, w/o UaL & 0.900 & 0.910 & 0.905   \\
\bottomrule
\end{tabular}}
\label{tab:ablation_results}
\end{table}


\begin{table*}
    \centering
    \caption{Results of knowledge erasure experiments. ``Rel.'', ``Gen.'' and ``Loc.'' represent reliability, generality, and locality metrics, respectively. ``average'' is the average value of the results on all language. The ``average'' of our method is followed by the improvements ($\uparrow$) over the previous state-of-the-art method AMIG.}
        \begin{tabular}{c|c|c|cccccccc}
        \toprule
        \textbf{Base Models} 
        & \textbf{Metrics}
        & \textbf{Methods} 
        & \textbf{en} 
        & \textbf{fr} 
        & \textbf{zh}
        & \textbf{fi} 
        & \textbf{hu} 
        & \textbf{ja} 
        & \textbf{ko} 
        & \textbf{average}
        \\
         \midrule
      \multirow{10}{*}{\textbf{mGPT}} 
         & \multirow{3}{*}{\textbf{Rel.}} 
         & Random & 39.56 & 24.03 & 35.62 & 40.50 & 34.05 & 35.92 & 29.41
& 34.16\\
    & &AMIG & 47.61 & 46.94 & 43.05 & 38.18 & 33.14 & 43.40 & 41.99
& 42.04\\
    & &  \textbf{\textsc{MaTrice} (Ours)} &  \textbf{51.00} & \textbf{59.48} & \textbf{55.18} & \textbf{59.92} & \textbf{58.35} & \textbf{56.31} & \textbf{59.09}  & \textbf{57.05} ($\uparrow$15.01) \\
         \cmidrule{2-11}
         & \multirow{3}{*}{\textbf{Gen.}} 
         & Random & 32.77 & 30.82 & 30.46 & 30.35 & 31.52 & 29.18 & 24.48
& 29.94\\
    & &AMIG & 35.62 & 40.48 & 32.97 & 30.00 & 40.78 & 43.77 & 37.02
& 37.23\\
    & &\textbf{\textsc{MaTrice} (Ours)} & \textbf{45.61} & \textbf{40.65} & \textbf{57.19} & \textbf{56.27} & \textbf{52.69} & \textbf{54.68} & \textbf{52.97}
& \textbf{51.44} ($\uparrow$14.21) \\
         \cmidrule{2-11}
         & \multirow{3}{*}{\textbf{Loc.}} 
         & Random & 6.56 & 6.78 & 5.42 & 6.38 & 4.37 & 6.80 & 5.24
& 5.94\\
    & &AMIG & \textbf{9.64} & 5.44 & \textbf{7.74} & \textbf{9.91} & 5.72 & \textbf{9.04} & 6.12
& 7.66\\
    & &\textbf{\textsc{MaTrice} (Ours)} & 9.19 & \textbf{9.92} & 7.26 & 7.10 & \textbf{10.65} & 5.86 & \textbf{9.26}
& \textbf{8.46} ($\uparrow$0.80) \\
         \midrule
         \midrule
         \multirow{10}{*}{\textbf{LLaMA2}} 
         & \multirow{3}{*}{\textbf{Rel.}} 
         & Random & 46.35 & 33.41 & 43.92 & 42.76 & 33.41 & 38.53 & 31.19
& 38.51\\
    & &AMIG & 49.39 & 54.77 & 51.73 & 40.49 & 37.76 & 47.27 & 50.40
& 47.40\\
    & &\textbf{\textsc{MaTrice} (Ours)} & \textbf{57.82} & \textbf{63.05} & \textbf{59.12} & \textbf{68.71} & \textbf{61.31} & \textbf{66.78} & \textbf{73.44}
& \textbf{64.32} ($\uparrow$16.92) \\
         \cmidrule{2-11}
         & \multirow{3}{*}{\textbf{Gen.}} 
         & Random & 36.89 & 33.15 & 43.17 & 35.50 & 32.18 & 29.65 & 25.75
& 33.76\\
    & &AMIG & 41.04 & 47.81 & 36.36 & 29.88 & 44.06 & 52.77 & 41.94
& 41.98\\
    & & \textbf{\textsc{MaTrice} (Ours)} & \textbf{57.38} & \textbf{50.51} & \textbf{56.30} & \textbf{64.24} & \textbf{60.63} & \textbf{60.51} & \textbf{56.38}
& \textbf{57.99} ($\uparrow$16.01) \\
         \cmidrule{2-11}
         & \multirow{3}{*}{\textbf{Loc.}} 
         & Random & 7.51 & 6.83 & 6.09 & 7.02 & 5.89 & 8.11 & 5.39
& 6.69\\
    & &AMIG & \textbf{10.51} & 5.68 & \textbf{9.48} & \textbf{11.54} & 6.02 & \textbf{10.10} & 7.13
& 8.64\\
    & & \textbf{\textsc{MaTrice} (Ours)} & 9.57 & \textbf{11.84} & 8.73 & 9.18 & \textbf{11.83} & 5.66 & \textbf{9.99}
& \textbf{9.54} ($\uparrow$0.90) \\
         \midrule
         \midrule
         \multirow{10}{*}{\textbf{LLaMA3}} 
         & \multirow{3}{*}{\textbf{Rel.}} 
         & Random & 38.96 & 36.28 & 46.86 & 51.19 & 37.11 & 35.78 & 39.80
& 40.85\\
    & &AMIG & \textbf{65.81} & 46.98 & 52.44 & 38.68 & 46.07 & 58.29 & 43.73
& 50.29\\
    & & \textbf{\textsc{MaTrice} (Ours)} & 61.87 & \textbf{72.49} & \textbf{61.54} & \textbf{81.17} & \textbf{67.52} & \textbf{61.77} & \textbf{71.27}
& \textbf{68.23} ($\uparrow$17.94) \\
         \cmidrule{2-11}
         & \multirow{3}{*}{\textbf{Gen.}} 
         & Random & 40.28 & 32.67 & 33.89 & 40.58 & 38.64 & 34.19 & 30.42
& 35.81\\
    & &AMIG & 38.52 & \textbf{46.89} & 42.72 & 47.68 & 47.43 & 49.65 & 38.85
& 44.53\\
    & & \textbf{\textsc{MaTrice} (Ours)} & \textbf{58.28} & 44.18 & \textbf{75.67} & \textbf{73.35} & \textbf{53.25} & \textbf{58.62} & \textbf{67.31}
& \textbf{61.52} ($\uparrow$16.99) \\
         \cmidrule{2-11}
         & \multirow{3}{*}{\textbf{Loc.}} 
         & Random & 6.95 & 6.95 & 7.52 & 7.59 & 5.41 & 8.47 & 6.81
& 7.10\\
    & &AMIG & 10.26 & 5.70 & \textbf{9.34} & \textbf{11.46} & 7.22 & \textbf{11.64} & 8.52
& 9.16\\
    & & \textbf{\textsc{MaTrice} (Ours)} & \textbf{13.16} & \textbf{10.09} & 8.86 & 8.37 & \textbf{13.85} & 6.68 & \textbf{9.86}
& \textbf{10.12} ($\uparrow$0.96)  \\
         \midrule
         \midrule
                  \multirow{10}{*}{\textbf{mBERT}} 
         & \multirow{3}{*}{\textbf{Rel.}} 
         & Random & 32.10 & 16.50 & 34.59 & 34.81 & 26.28 & 32.67 & 26.99
& 29.13\\
    & &AMIG & 40.57 & 42.24 & 38.60 & 29.64 & 26.08 & 40.88 & 33.01
& 35.86\\
    & & \textbf{\textsc{MaTrice} (Ours)} & \textbf{43.28} & \textbf{57.98} & \textbf{43.06} & \textbf{56.93} & \textbf{51.40} & \textbf{42.02} & \textbf{45.93}
& \textbf{48.66} ($\uparrow$12.80) \\
         \cmidrule{2-11}
         & \multirow{3}{*}{\textbf{Gen.}} 
         & Random & 31.51 & 24.90 & 25.31 & 23.07 & 28.51 & 22.00 & 23.45
& 25.54\\
    & &AMIG & 31.77 & \textbf{35.96} & 22.37 & 28.44 & 32.35 & 40.75 & 30.66
& 31.76\\
    & & \textbf{\textsc{MaTrice} (Ours)} & \textbf{36.96} & 30.12 & \textbf{41.29} & \textbf{47.03} & \textbf{49.32} & \textbf{51.09} & \textbf{51.30}
& \textbf{43.87} ($\uparrow$12.80) \\
         \cmidrule{2-11}
         & \multirow{3}{*}{\textbf{Loc.}} 
         & Random & 5.48 & 5.91 & 4.28 & 5.48 & 4.30 & 5.92 & 4.07
& 5.06\\
    & &AMIG & 8.17 & 4.39 & \textbf{7.01} & \textbf{8.97} & 3.93 & \textbf{8.33} & 4.94
& 6.53\\
    & & \textbf{\textsc{MaTrice} (Ours)} & \textbf{8.34} & \textbf{8.10} & 5.87 & 5.43 & \textbf{8.94} & 5.17 & \textbf{8.69}
& \textbf{7.22} ($\uparrow$0.69) \\
         \bottomrule
        \end{tabular}
    \label{erasure}
\end{table*}

\begin{table*}[!h]
    \centering
    \caption{Results of knowledge update experiment. ``en$\rightarrow$x'' represents that neuron editing is performed on the English data and then the edited model is evaluated on other languages data. Like knowledge erasure, we also adopt reliability, generality and locality as evaluation metrics.}
        \begin{tabular}{c|c|c|
        cccccccc}
        \toprule
        \textbf{Base Models} 
        & \textbf{Metrics}
        & \textbf{Methods} 
        & \textbf{en$\rightarrow$en} 
        & \textbf{en$\rightarrow$fr} 
        & \textbf{en$\rightarrow$zh}
        & \textbf{en$\rightarrow$fi} 
        & \textbf{en$\rightarrow$hu} 
        & \textbf{en$\rightarrow$ja} 
        & \textbf{en$\rightarrow$ko} 
        & \textbf{average}
        \\
         \midrule
         \multirow{10}{*}{\textbf{mGPT}} 
         & \multirow{3}{*}{\textbf{Rel.}} 
         & Random & 48.70 & 44.00 & 35.25  & 40.20 & 33.05 & 35.01  & 28.20
         & 37.77
         \\
         & & AMIG & 47.27 & 45.30& 42.93 & 37.33 & 31.18  & 43.30  & 41.22
         & 41.22
         \\
         & & \textbf{\textsc{MaTrice} (Ours)} & \textbf{69.60} & \textbf{58.30} & \textbf{54.70} & \textbf{58.65} & \textbf{56.60} & \textbf{56.30} & \textbf{57.35}
         & \textbf{58.79} ($\uparrow$17.57)
         \\
         \cmidrule{2-11}
         & \multirow{3}{*}{\textbf{Gen.}} 
         & Random & 41.70  & 40.30& 30.22  & 28.88 & 31.00 & 29.04  & 24.33
         & 32.21
         \\
         & & AMIG & 44.73 & 39.85 & 32.12 & 29.24  & 40.22 & 42.75 & 36.62
         & 37.93
         \\
         & & \textbf{\textsc{MaTrice} (Ours)} & \textbf{64.68} & \textbf{59.25} & \textbf{56.15} & \textbf{54.82}  & \textbf{51.72} & \textbf{53.49} & \textbf{52.89}
         & \textbf{56.14} ($\uparrow$18.21)
         \\
         \cmidrule{2-11}
         & \multirow{3}{*}{\textbf{Loc.}} 
         & Random  & 8.46 & 8.65 & 5.38 & 6.28 & 4.33 & 6.68 & 4.96
         & 6.39
         \\
         & & AMIG & 11.53 & 11.33 & \textbf{7.68} & \textbf{9.83} & 5.43 & \textbf{8.84} & 5.93
         &8.65
         \\
         & & \textbf{\textsc{MaTrice} (Ours)}& \textbf{16.14} & \textbf{12.69} & 7.21 & 6.98 & \textbf{10.33} & 5.66 & \textbf{9.08}
         & \textbf{9.73} ($\uparrow$1.08)
         \\
         \midrule
         \midrule
         \multirow{10}{*}{\textbf{LLaMA2}} 
         & \multirow{3}{*}{\textbf{Rel.}} 
         & Random & 56.01 & 44.29 & 38.40 & 42.75 & 41.47 & 42.75 & 38.40
& 43.44\\
    & &AMIG & 49.80 & 57.57 & 43.80 & 43.69 & 45.88 & 44.07 & 47.00
& 47.40\\
    & & \textbf{\textsc{MaTrice} (Ours)} & \textbf{80.66} & \textbf{67.58} & \textbf{58.45} & \textbf{68.68} & \textbf{70.38} & \textbf{56.41} & \textbf{71.07}
& \textbf{67.60} ($\uparrow$20.20) \\
         \cmidrule{2-11}
         & \multirow{3}{*}{\textbf{Gen.}} 
         & Random & 50.52 & 44.60 & 32.18 & 40.97 & 35.25 & 30.21 & 25.55
& 37.04\\
    & &AMIG & 51.65 & 44.78 & 38.71 & 35.20 & 44.60 & 50.70 & 39.71
& 43.62\\
    & & \textbf{\textsc{MaTrice} (Ours)} & \textbf{72.45} & \textbf{70.92} & \textbf{64.85} & \textbf{66.95} & \textbf{59.84} & \textbf{63.40} & \textbf{53.53}
& \textbf{64.56} ($\uparrow$20.94) \\
         \cmidrule{2-11}
         & \multirow{3}{*}{\textbf{Loc.}} 
         & Random & 9.29 & 9.70 & 5.67 & 7.13 & 4.70 & 7.69 & 7.27
& 7.35\\
    & &AMIG & 12.50 & 12.32 & \textbf{8.24} & \textbf{10.54} & 7.93 & \textbf{10.57} & 7.56
& 9.95\\
    & & \textbf{\textsc{MaTrice} (Ours)} & \textbf{16.62} & \textbf{14.74} & 7.67 & 8.05 & \textbf{13.11} & 7.46 & 10.65
& \textbf{11.19} ($\uparrow$1.24) \\
         \midrule
         \midrule
         \multirow{10}{*}{\textbf{LLaMA3}} 
         & \multirow{3}{*}{\textbf{Rel.}} 
         & Random & 61.35 & 59.58 & 49.59 & 44.43 & 33.64 & 40.84 & 33.15
& 46.08\\
    & &AMIG & 50.38 & 59.19 & 55.84 & 41.96 & 40.83 & 49.13 & 54.69
& 50.29\\
    & & \textbf{\textsc{MaTrice} (Ours)} & \textbf{82.28} & \textbf{65.62} & \textbf{61.51} & \textbf{74.16} & \textbf{63.86} & \textbf{72.45} & \textbf{82.15}
& \textbf{71.72} ($\uparrow$21.43)  \\
         \cmidrule{2-11}
         & \multirow{3}{*}{\textbf{Gen.}} 
         & Random & 50.05 & 44.88 & 51.07 & 39.53 & 32.90 & 30.02 & 26.62
& 39.30\\
    & &AMIG & 54.35 & 51.99 & 38.59 & 30.21 & 46.07 & 58.01 & 44.73
& 46.28\\
    & & \textbf{\textsc{MaTrice} (Ours)} & \textbf{85.41} & \textbf{77.63} & \textbf{56.40} & \textbf{70.21} & \textbf{66.27} & \textbf{64.95} & \textbf{58.59}
& \textbf{68.49} ($\uparrow$22.21)  \\
         \cmidrule{2-11}
         & \multirow{3}{*}{\textbf{Loc.}} 
         & Random & 10.15 & 8.94 & 6.53 & 7.48 & 6.84 & 8.99 & 5.65
& 7.80\\
    & &AMIG & 13.19 & 11.92 & \textbf{10.71} & \textbf{12.72} & 6.42 & \textbf{10.96} & 7.95
& 10.55\\
    & & \textbf{\textsc{MaTrice} (Ours)} & \textbf{16.88} & \textbf{16.39} & 9.83 & 10.76 & \textbf{12.90} & 5.66 & \textbf{10.65}
& \textbf{11.87} ($\uparrow$1.32) \\
         \midrule
         \midrule
                  \multirow{10}{*}{\textbf{mBERT}} 
         & \multirow{3}{*}{\textbf{Rel.}} 
         & Random & 48.53 & 35.82 & 27.51 & 32.87 & 30.34 & 34.50 & 20.47
& 32.86\\
    & &AMIG & 36.32 & 44.31 & 37.31 & 36.53 & 22.38 & 34.44 & 39.73
& 35.86\\
    & & \textbf{\textsc{MaTrice} (Ours)} & \textbf{61.98} & \textbf{49.49} & \textbf{50.45} & \textbf{44.66} & \textbf{49.82} & \textbf{52.90} & \textbf{48.70}
& \textbf{51.14} ($\uparrow$15.28) \\
         \cmidrule{2-11}
         & \multirow{3}{*}{\textbf{Gen.}} 
         & Random & 36.13 & 38.77 & 27.84 & 21.29 & 26.04 & 25.70 & 20.38
& 28.02\\
    & &AMIG & 42.40 & 35.53 & 25.61 & 17.92 & 35.79 & 38.51 & 35.25
& 33.00\\
    & & \textbf{\textsc{MaTrice} (Ours)} & \textbf{55.73} & \textbf{56.00} & \textbf{43.31} & \textbf{42.63} & \textbf{50.72} & \textbf{50.12} & \textbf{43.40}
& \textbf{48.84} ($\uparrow$15.84) \\
         \cmidrule{2-11}
         & \multirow{3}{*}{\textbf{Loc.}} 
             &Random & 8.14 & 8.45 & 3.99 & 5.43 & 3.63 & 5.52 & 3.76
& 5.56\\
    & &AMIG & 11.03 & 11.08 & \textbf{6.55} & \textbf{8.72} & 4.21 & \textbf{6.93} & 4.17
& 7.53\\
    & & \textbf{\textsc{MaTrice} (Ours)} & \textbf{13.35} & \textbf{12.41} & 6.07 & 6.02 & \textbf{7.89} & 4.96 & \textbf{8.54}
& \textbf{8.46} ($\uparrow$0.93) \\
         \bottomrule
        \end{tabular}
    \label{update}
\end{table*}


\begin{figure*}[ht]
        \centering
        \includegraphics[width=1.0\textwidth,height=0.9\textheight,keepaspectratio]{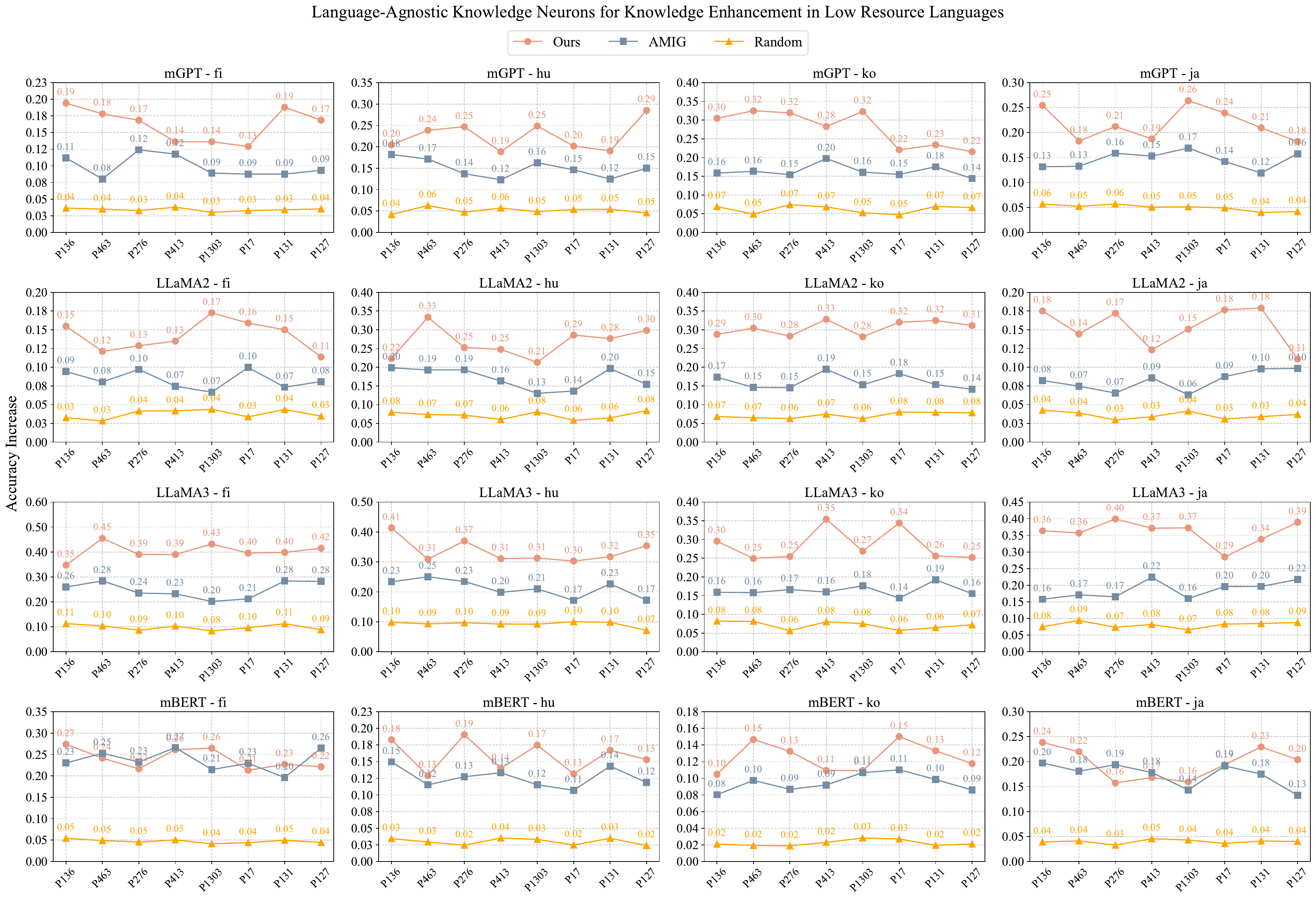}
    \caption{Results of knowledge enhancement experiment in low resource languages (i.e., Finnish (fi), Hungarian (hu), Korean (ko) and Japanese (ja)). We calculate the accuracy increase on the $Q_{\text{error}}$ by enhancing language-agnostic knowledge neurons.}
    \label{knowledge-enhancement}
\end{figure*}

\subsection{Language-Agnostic Knowledge Neurons for Cross-Lingual Knowledge Editing} \label{KE}
In this subsection, we conduct cross-lingual knowledge editing experiments to explore the potential applications of language-agnostic knowledge neurons (\textbf{RQ3}).

\textbf{Experimental Settings:} Given a fact $\langle$$s, r, o$$\rangle$, we perform two types of knowledge modification: Erasure and Update, which corresponds to different editing ways of the FFN weights:
\begin{equation}
\label{eq-edit}
\bm{W}_{i}^{l} = \begin{cases}
    \bm{0}, & \text{if Erasure} \\
    \bm{W}_{i}^{l} - \lambda_1\bm{E}(o) + \lambda_2\bm{E}(o'), & \text{if Update}
\end{cases}
\end{equation}
where $\bm{W}_{i}^{l}$ is the FFN weight corresponding to the language-agnostic knowledge neuron $n_{i}^{l}$. $\bm{E}(o)$ and $\bm{E}(o')$ are embeddings of the original object $o$ and the updated object $o'$, respectively. $\lambda_1$ and $\lambda_2$ are hyperparameters. For knowledge erasure, we set the FFN weights to zero vectors. For knowledge update, we use the embeddings of objects expressed in English to modify the FFN weights. 

\textit{Evaluation Metrics}: In cross-lingual editing settings, we edit the model in source language $l_{s}$ (i.e., English), and evaluate the edited model in target language $l_{t}$. We adopt the reliability, generality, and locality as evaluation metrics:

1) \textit{Reliability} measures whether the examples edited by the language $l_{s}$ can be expressed by the target language $l_{t}$:
\begin{equation}
\mathbb{E}_{(x_e, y_e) \in S_{e}(l_t)} \mathbbm{1} \left\{\operatorname{argmax}_y f_{\theta_{e}}\left(y | x_{e}\right) = y_{e}\right\},
\end{equation}
where $f_{\theta_{e}}(\cdot)$ denotes the edited model. $S_{e}(l_t)$ is the collection of all edited examples in the target language $l_t$. $\mathbbm{1}\{\cdot\}$ is the indicator function.

2) \textit{Generality} refers to the ability of the edited model to generate consistent outputs for semantically equivalent queries: 
\begin{equation}
\mathbb{E}_{(x_e, y_e) \in S_{p}(l_t)} \mathbbm{1} \left\{\operatorname{argmax}_y f_{\theta_{e}}\left(y| x_{e}\right) = y_{e}\right\},
\end{equation}
where $S_{p}(l_t)$ denotes the collection of all paraphrased examples in the target language $l_t$.

3) \textit{Locality} assesses whether editing new knowledge for LLMs will have a negative impact on the expression of irrelevant knowledge:
\begin{equation}
\mathbb{E}_{(x_e, y_e) \in S_{n}(l_t)} \mathbbm{1} \left\{\operatorname{argmax}_y f_{\theta_{e}}\left(y| x_{e}\right) = f_{\theta}\left(y|x_{e}\right) \right\},
\end{equation}
where $S_{n}(l_t)$ is the collection of examples unrelated to the edited knowledge in the target language $l_t$. $f_{\theta}(\cdot)$ denotes the model before editing. 

\textbf{Results and Analysis:} Table \ref{erasure} and Table \ref{update} present the results of knowledge erasure and update, respectively, from which we can draw the following key conclusions: 

1) Our method outperforms previous approaches for all evaluation metrics by a large margin. For example, compared with the AMIG method based on the mGPT, our method achieves improvements of 15.01\% for knowledge erasure and 17.57\% for knowledge update in terms of average reliability score. The substantial performance gain of our method over baselines demonstrates that the proposed \textsc{MaTrice} is very effective to localize language-agnostic knowledge neurons. Besides, our method and the AMIG both have significantly better performance than the Random method, which suggests that language-agnostic knowledge neurons have the potential to be applied in cross-lingual knowledge editing.

2) Compared to reliability and generality scores, all the three methods have lower locality score on the four LLMs. 
The reason is that if we desire to improve the reliability and generality scores in multiple languages, each method will have to edit a relatively large number of neurons. For instance, when using the threshold method to select neurons, a lower threshold is employed, which leads to noise and damage to other factual knowledge.
In fact, this is also a bottleneck in knowledge editing, which is how to avoid affecting other irrelevant knowledge during the editing process. Therefore, more effective editing methods are expected to be proposed to improve locality scores.

\begin{table*}[t!]
\centering
\caption{Results of the learning new knowledge experiment. ``Avg'' is the average of performance on ``$Q_{\text{new}}$'' and ``$Q_{\text{old}}$''. The ``Avg'' of our method is followed by the improvements ($\uparrow$) over the previous state-of-the-art method amig\_ft.}
\label{new-knowledge}
\scalebox{0.91}{
\begin{tabular}
{c
ccc
ccc
ccc
ccc}
\toprule
\multirow{4}{*}{\textbf{Methods}} & \multicolumn{12}{c}{\textbf{English (en)}} \\
\cmidrule{2-13}
\rowcolor{gray!10} & \multicolumn{3}{c}{\cellcolor{blue!10}mGPT} & \multicolumn{3}{c}{\cellcolor{green!10}LLaMA2} & \multicolumn{3}{c}{\cellcolor{yellow!10}LLaMA3} & \multicolumn{3}{c}{\cellcolor{red!10}mBERT}  \\
\cmidrule(lr){2-4}
\cmidrule(lr){5-7}
\cmidrule(lr){8-10}
\cmidrule(lr){11-13}
& $Q_{\text{new}}$ & $Q_{\text{old}}$ & Avg & $Q_{\text{new}}$ & $Q_{\text{old}}$ & Avg & $Q_{\text{new}}$ & $Q_{\text{old}}$ & Avg & $Q_{\text{new}}$ & $Q_{\text{old}}$ & Avg \\
\midrule
Direct\_FT & \textbf{98.12} & 61.23 & 80.18 & \textbf{100} & 72.35 & 86.18 & \textbf{100} & 75.42 & 88.21 & \textbf{96.31} & 55.19 & 76.25 \\
Random\_FT & 57.38 & 64.27 & 61.33 & 62.15 & 70.48 & 66.32 & 64.29 & 72.17 & 68.23 & 52.41 & 60.36 & 56.39 \\
AMIG\_FT & 75.26 & 76.43 & 76.35 & 80.19 & 81.32 & 81.26 & 81.47 & 79.28 & 80.38 & 72.34 & 71.29 & 72.32 \\
\textbf{\textsc{MaTrice} (Ours)} & 94.37 & \textbf{87.21} & \textbf{91.29} ($\uparrow$14.94) & 97.43 & \textbf{92.18} & \textbf{95.31} ($\uparrow$14.05) & 98.26 & \textbf{86.39} & \textbf{92.33} ($\uparrow$11.95) & 92.45 & \textbf{82.37} & \textbf{87.41} ($\uparrow$15.09) \\
\midrule
\multirow{4}{*}{\textbf{Methods}} & \multicolumn{12}{c}{\textbf{French (fr)}} \\
\cmidrule{2-13}
& \multicolumn{3}{c}{\cellcolor{blue!10}mGPT} & \multicolumn{3}{c}{\cellcolor{green!10}LLaMA2} & \multicolumn{3}{c}{\cellcolor{yellow!10}LLaMA3} & \multicolumn{3}{c}{\cellcolor{red!10}mBERT}  \\
\cmidrule(lr){2-4}
\cmidrule(lr){5-7}
\cmidrule(lr){8-10}
\cmidrule(lr){11-13}
& $Q_{\text{new}}$ & $Q_{\text{old}}$ & Avg & $Q_{\text{new}}$ & $Q_{\text{old}}$ & Avg & $Q_{\text{new}}$ & $Q_{\text{old}}$ & Avg & $Q_{\text{new}}$ & $Q_{\text{old}}$ & Avg \\
\midrule
Direct\_FT & 58.27 & 49.35 & 54.31 & 60.18 & 58.42 & 59.30 & 60.39 & 60.24 & 60.32 & 57.16 & 44.28 & 51.22  \\
Random\_FT & 46.29 & 51.37 & 49.33 & 50.21 & 56.38 & 53.30 & 51.42 & 58.19 & 55.31 & 42.35 & 48.27 & 45.31 \\
AMIG\_FT & 60.31 & 61.24 & 61.28 & 64.37 & 65.19 & 65.28 & 65.42 & 63.31 & 64.37 & 58.29 & 57.36 & 58.33 \\
\textbf{\textsc{MaTrice} (Ours)}& \textbf{75.38} & \textbf{70.27} & \textbf{73.33} ($\uparrow$12.05) & \textbf{78.41} & \textbf{74.35} & \textbf{76.38} ($\uparrow$11.10) & \textbf{78.29} & \textbf{69.37} & \textbf{74.33} ($\uparrow$9.96) & \textbf{74.42} & \textbf{66.31} & \textbf{70.37} ($\uparrow$12.04) \\
\midrule
\multirow{4}{*}{\textbf{Methods}} & \multicolumn{12}{c}{\textbf{Finnish (fi)}} \\
\cmidrule{2-13}
\rowcolor{gray!10} & \multicolumn{3}{c}{\cellcolor{blue!10}mGPT} & \multicolumn{3}{c}{\cellcolor{green!10}LLaMA2} & \multicolumn{3}{c}{\cellcolor{yellow!10}LLaMA3} & \multicolumn{3}{c}{\cellcolor{red!10}mBERT}  \\
\cmidrule(lr){2-4}
\cmidrule(lr){5-7}
\cmidrule(lr){8-10}
\cmidrule(lr){11-13}
& $Q_{\text{new}}$ & $Q_{\text{old}}$ & Avg & $Q_{\text{new}}$ & $Q_{\text{old}}$ & Avg & $Q_{\text{new}}$ & $Q_{\text{old}}$ & Avg & $Q_{\text{new}}$ & $Q_{\text{old}}$ & Avg \\
\midrule
Direct\_FT & 39.28 & 37.41 & 38.35 & 40.19 & 43.27 & 42.23 & 40.36 & 45.28 & 43.32 & 38.24 & 33.37 & 36.31 \\
Random\_FT & 34.32 & 38.29 & 36.31 & 37.41 & 42.35 & 40.38 & 38.27 & 43.39 & 41.33 & 31.42 & 36.28 & 34.35 \\
AMIG\_FT & 45.29 & 46.37 & 46.33 & 48.41 & 49.26 & 49.34 & 49.38 & 47.32 & 48.35 & 43.27 & 43.39 & 43.33 \\
\textbf{\textsc{MaTrice} (Ours)} & \textbf{56.41} & \textbf{52.38} & \textbf{54.40} ($\uparrow$8.07) & \textbf{58.27} & \textbf{55.39} & \textbf{57.33} ($\uparrow$7.99) & \textbf{59.32} & \textbf{52.28} & \textbf{56.30} ($\uparrow$7.95) & \textbf{55.36} & \textbf{49.41} & \textbf{52.39} ($\uparrow$9.06) \\
\midrule
\multirow{4}{*}{\textbf{Methods}} & \multicolumn{12}{c}{\textbf{Hungarian (hu)}} \\
\cmidrule{2-13}
\rowcolor{gray!10} & \multicolumn{3}{c}{\cellcolor{blue!10}mGPT} & \multicolumn{3}{c}{\cellcolor{green!10}LLaMA2} & \multicolumn{3}{c}{\cellcolor{yellow!10}LLaMA3} & \multicolumn{3}{c}{\cellcolor{red!10}mBERT}  \\
\cmidrule(lr){2-4}
\cmidrule(lr){5-7}
\cmidrule(lr){8-10}
\cmidrule(lr){11-13}
& $Q_{\text{new}}$ & $Q_{\text{old}}$ & Avg & $Q_{\text{new}}$ & $Q_{\text{old}}$ & Avg & $Q_{\text{new}}$ & $Q_{\text{old}}$ & Avg & $Q_{\text{new}}$ & $Q_{\text{old}}$ & Avg \\
\midrule
Direct\_FT & 29.36 & 31.24 & 30.30 & 30.42 & 36.27 & 33.35 & 30.39 & 38.21 & 34.30 & 28.33 & 28.19 & 28.26 \\
Random\_FT & 29.28 & 32.37 & 31.33 & 31.41 & 35.29 & 33.35 & 32.38 & 36.27 & 34.33 & 26.42 & 30.31 & 28.37 \\
AMIG\_FT & 38.32 & 38.41 & 38.37 & 40.28 & 41.36 & 41.32 & 41.39 & 40.27 & 41.33 & 36.41 & 36.29 & 36.35 \\
 \textbf{\textsc{MaTrice} (Ours)} & \textbf{47.39} & \textbf{44.28} & \textbf{46.34} ($\uparrow$7.97) & \textbf{49.41} & \textbf{46.37} & \textbf{48.39} ($\uparrow$7.07) & \textbf{49.32} & \textbf{43.28} & \textbf{46.30} ($\uparrow$4.97) & \textbf{46.37} & \textbf{41.42} & \textbf{44.40} ($\uparrow$8.05) \\
\midrule
\multirow{4}{*}{\textbf{Methods}} & \multicolumn{12}{c}{\textbf{Japanese (ja)}} \\
\cmidrule{2-13}
\rowcolor{gray!10} & \multicolumn{3}{c}{\cellcolor{blue!10}mGPT} & \multicolumn{3}{c}{\cellcolor{green!10}LLaMA2} & \multicolumn{3}{c}{\cellcolor{yellow!10}LLaMA3} & \multicolumn{3}{c}{\cellcolor{red!10}mBERT}  \\
\cmidrule(lr){2-4}
\cmidrule(lr){5-7}
\cmidrule(lr){8-10}
\cmidrule(lr){11-13}
& $Q_{\text{new}}$ & $Q_{\text{old}}$ & Avg & $Q_{\text{new}}$ & $Q_{\text{old}}$ & Avg & $Q_{\text{new}}$ & $Q_{\text{old}}$ & Avg & $Q_{\text{new}}$ & $Q_{\text{old}}$ & Avg \\
\midrule
Direct\_FT & 34.29 & 34.37 & 34.33 & 35.41 & 40.28 & 38.35 & 35.36 & 41.29 & 38.33 & 33.42 & 30.31 & 32.37 \\
Random\_FT & 31.38 & 35.27 & 33.33 & 34.41 & 39.32 & 37.37 & 35.28 & 40.39 & 38.34 & 29.36 & 33.41 & 31.39 \\
AMIG\_FT & 41.29 & 42.37 & 42.33 & 44.41 & 45.28 & 45.35 & 45.39 & 43.32 & 44.36 & 40.27 & 39.38 & 40.33 \\
\textbf{\textsc{MaTrice} (Ours)} & \textbf{52.38} & \textbf{48.41} & \textbf{50.40} ($\uparrow$8.07) & \textbf{53.29} & \textbf{51.37} & \textbf{52.33} ($\uparrow$6.98) & \textbf{54.32} & \textbf{47.28} & \textbf{51.30} ($\uparrow$6.94)  & \textbf{51.36} & \textbf{45.41} & \textbf{48.39} ($\uparrow$8.06) \\
\midrule
\multirow{4}{*}{\textbf{Methods}} & \multicolumn{12}{c}{\textbf{Korean (ko)}} \\
\cmidrule{2-13}
\rowcolor{gray!10} & \multicolumn{3}{c}{\cellcolor{blue!10}mGPT} & \multicolumn{3}{c}{\cellcolor{green!10}LLaMA2} & \multicolumn{3}{c}{\cellcolor{yellow!10}LLaMA3} & \multicolumn{3}{c}{\cellcolor{red!10}mBERT}  \\
\cmidrule(lr){2-4}
\cmidrule(lr){5-7}
\cmidrule(lr){8-10}
\cmidrule(lr){11-13}
& $Q_{\text{new}}$ & $Q_{\text{old}}$ & Avg & $Q_{\text{new}}$ & $Q_{\text{old}}$ & Avg & $Q_{\text{new}}$ & $Q_{\text{old}}$ & Avg & $Q_{\text{new}}$ & $Q_{\text{old}}$ & Avg \\
\midrule
Direct\_FT  & 24.38 & 27.29 & 26.34 & 25.41 & 32.27 & 29.34 & 25.36 & 34.28 & 30.32 & 23.42 & 25.31 & 24.37  \\
Random\_FT & 26.31 & 29.38 & 28.35 & 28.42 & 32.29 & 30.36 & 29.37 & 33.41 & 31.39 & 23.28 & 27.36 & 25.32 \\
AMIG\_FT & 34.39 & 34.27 & 34.33 & 36.41 & 36.28 & 36.35 & 36.38 & 35.29 & 36.34 & 32.42 & 32.31 & 32.37 \\
\textbf{\textsc{MaTrice} (Ours)} & \textbf{42.37} & \textbf{39.41} & \textbf{41.39} ($\uparrow$7.06) & \textbf{44.28} & \textbf{41.36} & \textbf{43.32} ($\uparrow$6.97) & \textbf{44.39} & \textbf{39.27} & \textbf{42.33} ($\uparrow$5.99) & \textbf{41.41} & \textbf{37.29} & \textbf{39.35} ($\uparrow$6.98) \\
\midrule
\multirow{4}{*}{\textbf{Methods}} & \multicolumn{12}{c}{\textbf{Chinese (zh)}} \\
\cmidrule{2-13}
& \multicolumn{3}{c}{mGPT} & \multicolumn{3}{c}{LLaMA2} & \multicolumn{3}{c}{LLaMA3} & \multicolumn{3}{c}{mBERT}  \\
\cmidrule(lr){2-4}
\cmidrule(lr){5-7}
\cmidrule(lr){8-10}
\cmidrule(lr){11-13}
& $Q_{\text{new}}$ & $Q_{\text{old}}$ & Avg & $Q_{\text{new}}$ & $Q_{\text{old}}$ & Avg & $Q_{\text{new}}$ & $Q_{\text{old}}$ & Avg & $Q_{\text{new}}$ & $Q_{\text{old}}$ & Avg \\
\midrule
Direct\_FT & 44.32 & 40.28 & 42.30 & 45.39 & 47.41 & 46.40 & 45.36 & 49.27 & 47.32 & 42.41 & 36.29 & 39.35  \\
Random\_FT & 37.38 & 41.29 & 39.34 & 40.41 & 45.27 & 43.34 & 41.36 & 46.38 & 44.37 & 34.42 & 39.31 & 37.37 \\
AMIG\_FT & 49.29 & 49.37 & 49.33 & 52.41 & 52.28 & 52.35 & 53.38 & 51.32 & 52.35 & 47.27 & 46.39 & 47.33 \\
\textbf{\textsc{MaTrice} (Ours)} & \textbf{61.38} & \textbf{57.41} & \textbf{59.40} ($\uparrow$10.07) & \textbf{63.29} & \textbf{60.37} & \textbf{62.33} ($\uparrow$9.98) & \textbf{64.32} & \textbf{56.28} & \textbf{60.30} ($\uparrow$7.95) & \textbf{60.36} & \textbf{54.41} & \textbf{57.39} ($\uparrow$10.06) \\
\bottomrule
\end{tabular}}
\end{table*}

\subsection{Language-Agnostic Knowledge Neurons for Knowledge Enhancement in Low Resource Languages} \label{KELRL}
In this subsection, we investigate whether language-agnostic knowledge neurons are helpful for enhancing the knowledge expression of LLMs in low resource languages (\textbf{RQ3}). 

\textbf{Experimental Settings:} We select Finnish (fi), Hungarian (hu), Korean (ko) and Japanese (ja) as low resource languages, because the mastery of these four languages by LLMs is not as good as that of English, Chinese and French. On these four low resource languages, we filter the queries from the new \texttt{RML-LAMA} dataset and obtain some queries that LLMs originally answer incorrectly, denoted as $Q_{\text{error}}$. Thus, the original accuracy of LLMs on the dataset $Q_{\text{error}}$ is exactly 0. 
Then, we perform the enhancement operation on language-agnostic knowledge neurons corresponding to these queries, and calculate the increase in accuracy of modified LLMs on the $Q_{\text{error}}$.

\textbf{Results and Analysis:} Figure \ref{knowledge-enhancement} shows the results of knowledge enhancement experiment in four low resource languages. From the figure, we can observe that:

1) Our method and the AMIG both outperform the Random method on the four LLMs, which indicates that utilizing language-agnostic knowledge neurons helps improve performance of LLMs under low resource languages. 

2) Our method achieves better performance than the AMIG. In detail, for the Finnish (fi), our method based on LLaMA3 improves accuracy by 45\% for the P463 relation, while the AMIG only achieves an accuracy improvement of 28\%. This demonstrate that our method localizes language-agnostic knowledge neurons more accurately.

3) As the capability of LLMs becomes stronger (i.e., mBERT< mGPT<LLaMA2<LLaMA3), the performance improvement brought by our method is gradually increasing. The reason may be that neurons in larger models contain richer knowledge and language features.

\subsection{Language-Agnostic Knowledge Neurons Guide LLMs to Learn New Knowledge} \label{LEK}
In this subsection, we investigate the role of language-agnostic knowledge neurons in learning new knowledge (\textbf{RQ3}). 

\textbf{Experimental Settings:} We first identify language-agnostic knowledge neurons based on the complete \texttt{RML-LAMA} dataset. From this dataset, we then derive $Q_{\text{new}}$, a subset of queries introducing new knowledge that LLMs initially answer incorrectly. Using $Q_{\text{new}}$, we only fine-tune the parameters associated with language-agnostic knowledge neurons to inject the new knowledge into LLMs, while freezing all other parameters during training. 

\textit{Baseline}: Our method, \textsc{MaTrice}, is compared against three baselines: 1) \textbf{Direct\_FT}, which directly fine-tunes all parameters of LLMs; 2) \textbf{Random\_FT}, which randomly selects a set of neurons equal in number to those obtained by our method \textsc{MaTrice}, and then update the corresponding parameters; and 3) \textbf{AMIG\_FT}, which only fine-tunes the parameters associated language-agnostic knowledge neurons identified by the AMIG method. 

To evaluate the effectiveness of our approach, we assess the updated models' performance on two datasets: 1) $Q_{\text{new}}$, to measure the LLMs' acquisition of the newly introduced knowledge, and 2) $Q_{\text{old}}$, which contains the queries that LLMs can answer correctly, to determine if the LLMs retain the old knowledge during learning new knowledge.

\textbf{Results and Analysis:} Table \ref{new-knowledge} presents the result (i.e., prediction accuracy) of learning new knowledge experiment. From the results, we have two important observations: 

1) Our method \textsc{MaTrice} and the AMIG\_FT obtain higher average score by only fine-tuning language-agnostic knowledge neurons under all settings, which demonstrates that language-agnostic knowledge neurons enable LLMs to learn new knowledge more efficiently while preserving old knowledge simultaneously. 

2) Our method \textsc{MaTrice} can achieve better performance than the AMIG\_FT. For example, for the English, our method \textsc{MaTrice} can achieve an average performance (i.e., Avg) improvement of 14.05\% and 11.95\% based on LLaMA2 and LLaMA3, respectively. It indicates that our method can localize more precise language-agnostic knowledge neurons. The experiment may offer a potential solution to the problem of catastrophic forgetting during learning new knowledge.



\subsection{Case Study about Knowledge Enhancement and Knowledge Update} \label{case study}
\begin{figure}
        \centering
        \includegraphics[width=\linewidth]{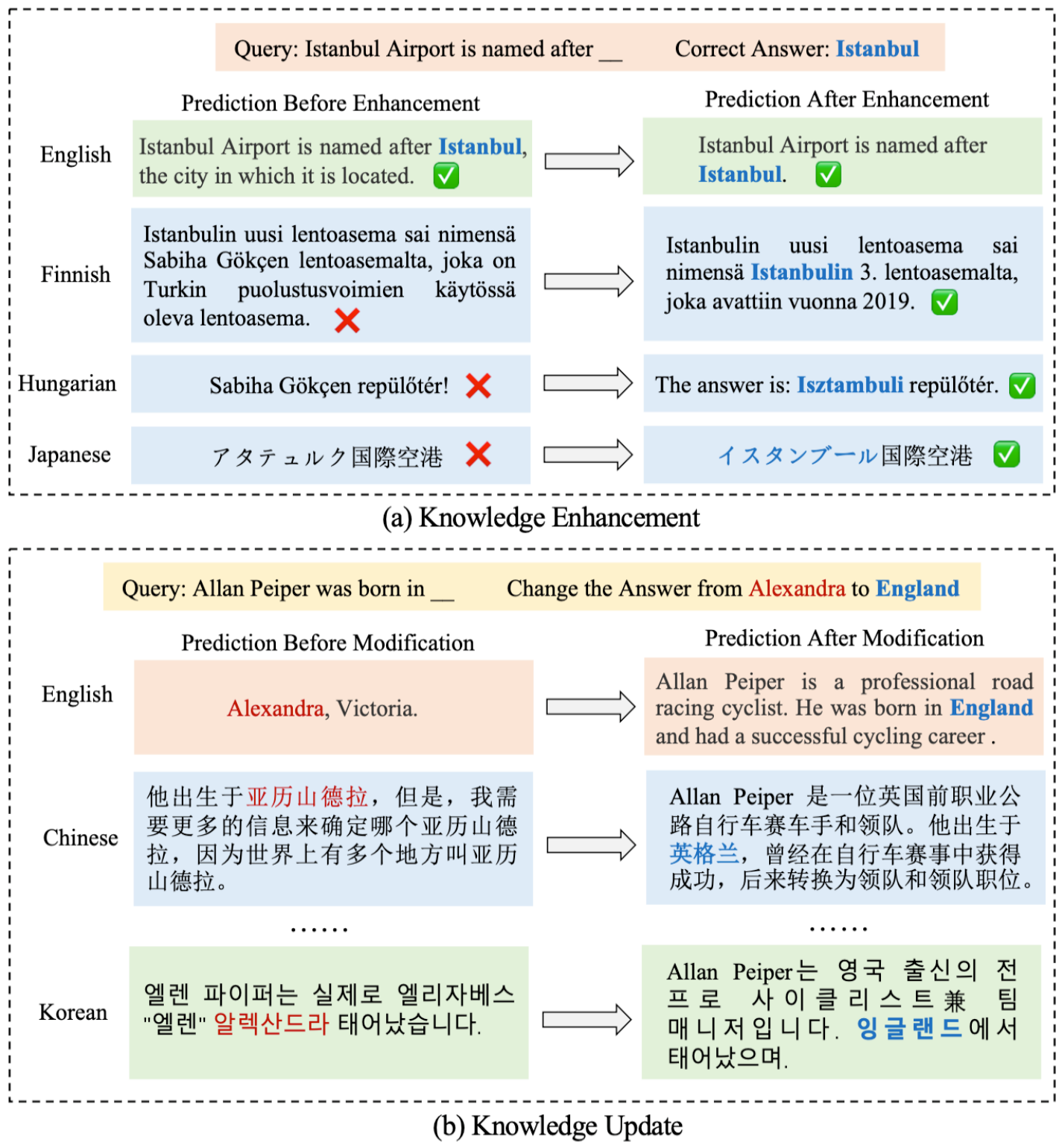}
\caption{Two examples to qualitatively illustrate the effectiveness of the language-agnostic knowledge neurons for knowledge enhancement and knowledge update. (a) Knowledge Enhancement: Given the query ``\textit{Istanbul Airport is named after\_\_}'', the model can give correct prediction (i.e., \textit{Istanbul} ) in low resource languages by enhancing language-agnostic knowledge neurons. (b) Knowledge Update: Given the query ``\textit{Allan Peiper was born in \_\_}'', we can change the prediction from \textit{Alexandra} to \textit{England} for multiple languages by modifying language-agnostic knowledge neurons. In the case study, we adopt the LLaMA3 as the base model.}
    \label{case}
\end{figure}

We give an example shown in Figure \ref{case}(a) to illustrate the effectiveness of using language-agnostic knowledge neurons for enhancing the knowledge expression of LLMs in low resource languages. For the cloze-style query ``\textit{Istanbul Airport is named after} \_\_'', LLaMA3 cannot make accurate predictions under low resource languages, including Finnish, Hungarian, Japanese and Korean. For the high resource language (i.e., English), LLaMA3 can provide the correct answer. We perform enhancement operations on language-agnostic knowledge neurons and find that the enhanced model can provide correct answers for both low resource and high resource languages. It shows that enhancing language-agnostic knowledge neurons enables to improve knowledge expression in low resource languages.

Besides, as shown in Figure \ref{case}(b), for the cloze-style query ``\textit{Allan Peiper was born in} \_\_'', we modify the language-agnostic knowledge neurons to change the answer from \textit{Alexandra} to \textit{England}. We observe that the edited model can successfully provides the target answer (i.e., \textit{England}) for all languages. This demonstrates that language-agnostic knowledge neurons contribute to cross-lingual knowledge editing.

\section{Conclusion and Future Work}

In this paper, to accurately analyze multilingual knowledge storage mechanisms in LLMs, we first construct a high-quality cloze-style multilingual benchmark called \texttt{RML-LAMA}, and then propose a novel method named multilingual integrated gradients with uncertainty estimation (\textsc{MaTrice}) to localize language-agnostic knowledge neurons. We visualize the distribution of language-agnostic knowledge neurons in LLMs with different architectures, and prove the accuracy of positioning results through neuron manipulation experiments. Furthermore, we explore the potential role of language-agnostic knowledge neurons in cross-lingual knowledge editing, knowledge enhancement and knowledge injection.

Future work could extend our approach, which is currently based on knowledge neuron theory \cite{geva2021transformer} and focuses on factual knowledge, to investigate other types of knowledge, such as commonsense knowledge and event knowledge. Additionally, as the theory of knowledge neurons evolves, our method can be adapted to new theoretical frameworks, potentially leading to more significant insights.

\bibliographystyle{IEEEtran}
\bibliography{IEEEabrv,paper}

\end{document}